\documentclass[sigconf,table]{acmart} % table,prologue options are needed for xcolor package
\usepackage{graphicx}
\usepackage{caption}
\usepackage{subcaption}
\usepackage{adjustbox}
\usepackage{xspace}
\usepackage[most]{tcolorbox}
\usepackage{enumitem}
\usepackage{fontawesome5}
\usepackage{booktabs}
\usepackage{multirow}
\usepackage[frozencache,cachedir=minted-cache]{minted} % for arXiv

\newcommand\vldbavailabilityurl{https://github.com/VIDA-NYU/harmonia}
\copyrightyear{2025}
\acmYear{2025}
\setcopyright{acmlicensed}\acmConference[NOVAS '25]{Novel Optimizations for Visionary AI Systems}{June 22--27, 2025}{Berlin, Germany}
\acmBooktitle{Novel Optimizations for Visionary AI Systems (NOVAS '25), June 22--27, 2025, Berlin, Germany}
\acmDOI{10.1145/3735079.3735324}
\acmISBN{979-8-4007-1917-2/2025/06}

\def\withnotes{0} % change to (0) to hide all comments, (1) to show all comments
\def\withrevisionhighlight{0}

\setlist[itemize]{leftmargin=4mm} % requires package enumitem

\ifnum\withnotes=1    
    \definecolor{Maroon}{rgb}{0.62, 0.0, 0.09}
    \definecolor{Emerald}{rgb}{.07, .74, .62}
    \definecolor{Orange}{rgb}{1.0, 0.55, 0.2}
    \newcommand\todo[1]{
        \textcolor{blue}{\textbf{TODO:} #1}
    }
    \newcommand{\as}[1]{\textcolor{violet}{\textbf{Aécio:} #1}}
    \newcommand{\jf}[1]{\textcolor{Emerald}{\textbf{Juliana:} #1}}
    \newcommand{\ep}[1]{\textcolor{Orange}{\textbf{Eduardo:} #1}}
    \newcommand{\rl}[1]{\textcolor{cyan}{\textbf{Roque:} #1}}
\else
    \newcommand{\asnote}[1]{}
    \newcommand{\as}[1]{}
    \newcommand{\jf}[1]{}
    \newcommand{\ep}[1]{}
    \newcommand{\rl}[1]{}
    \newcommand{\todo}[1]{}
\fi

\definecolor{HighlightColor}{rgb}{0.05,0.05,0.70}
\newcommand{\revised}[1]{{\color{HighlightColor}#1}}
\ifnum\withrevisionhighlight=0
    \renewcommand{\revised}[1]{#1}
\fi

\newcommand{\ignore}[1]{\leavevmode\unskip} % eat unnecessary spaces before

\newcommand{\myparagraph}[1]{\vspace{0.15em}\noindent \textbf{#1.}}

\theoremstyle{definition}

\newtheorem{definition}{Definition}

\newtheorem{example}{Example}

\newcommand{\hide}[1]{}
\newcommand{\hidecomment}[1]{}

\newtcblisting{qwr}{listing only,colback=teal!2.5!white,colframe=teal!85!black,boxrule=0.25mm,boxsep=4pt,left=0pt,right=0pt,top=0pt,bottom=0pt}

\newcommand{\chatboxAgent}[1]{
    \begin{tcolorbox}[colback=teal!2.5!white,colframe=teal!85!black,boxrule=0.25mm,boxsep=4pt,left=0pt,right=0pt,top=0pt,bottom=0pt]
    \scriptsize
    \tt
    { \scriptsize \faRobot }
    #1
    \end{tcolorbox}
}

\newcommand{\chatboxUser}[1]{
    \begin{tcolorbox}[colback=violet!2.5!white,colframe=violet!85!black,boxrule=0.25mm,boxsep=4pt,left=0pt,right=0pt,top=0pt,bottom=0pt]
    \scriptsize
    \tt
    { \scriptsize \faUser }
    #1
    \end{tcolorbox}
}

\newtcolorbox{UserChatbox}[1]{
colback=violet!2.5!white,
colframe=violet!85!black,
boxrule=0.25mm,
boxsep=4pt,
left=0pt,
right=0pt,
top=0pt,
bottom=0pt
}

\begin{document}

% \title{Interactive Data Harmonization with LLM Agents [Vision]}
\title[
Interactive Data Harmonization with LLM Agents: Opportunities and Challenges % short title used in page headers.
]{
Interactive Data Harmonization with LLM Agents: \\ Opportunities and Challenges
}

% \author{Aécio Santos}
% \affiliation{%
% 	\institution{New York University}
% 	\city{}
% 	\state{}
% }
% \email{aecio.santos@nyu.edu}

% \author{Eduardo H. M. Pena}
% \affiliation{%
% 	\institution{Federal University of Technology - Paran\'{a}}
% 	\city{}
% 	\state{}
% }
% \email{eduardopena@utfpr.edu.br}
% \authornote{Work done as a visiting researcher at New York University.}

% \author{Roque Lopez}
% \affiliation{%
% 	\institution{New York University}
% 	\city{}
% 	\state{}
% }
% \email{rlopez@nyu.edu}

% \author{Juliana Freire}
% \affiliation{%
% 	\institution{New York University}
% 	\city{}
% 	\state{}
% }
% \email{juliana.freire@nyu.edu}

% Author list
\author{Aécio Santos$^1$, \ Eduardo H. M. Pena$^2$, \ Roque Lopez$^1$, \ Juliana Freire$^1$}
\affiliation{
  \institution{$^1$New York University}
}
\email{{aecio.santos,rlopez,juliana.freire}@nyu.edu}
\affiliation{
  \institution{$^2$Federal University of Technology - Paraná}
}
\email{eduardopena@utfpr.edu.br}
% \authornote{Work done while at New York University.}
% \authornotemark[2]

\renewcommand{\shortauthors}{Aécio Santos, \  Eduardo H. M. Pena, \ Roque Lopez, \ Juliana Freire}
\renewcommand{\authors}{Aécio Santos, \  Eduardo H. M. Pena, \ Roque Lopez, \ Juliana Freire}

\newcommand{\systemname}{\texttt{Harmonia}\xspace}
\newcommand{\bdikit}{\texttt{bdi-kit}\xspace}

\begin{abstract}
% Data harmonization is essential for integrating datasets from diverse sources, yet it remains a time-consuming and challenging task due to schema mismatches, varying terminologies, and differences in data collection methodologies.
% This paper outlines our vision for agentic data harmonization systems and introduces \systemname, a novel system that embodies this vision. 
% To automate the synthesis of data harmonization pipelines, our approach combines LLM-based reasoning with a library of established and efficient data integration algorithms that address common data integration challenges. Together with domain experts, these components enable the interactive development of data harmonization pipelines that can be later reused to recreate harmonized datasets.
% We demonstrate \systemname's application in a real-world scenario involving clinical data integration and mapping datasets to standardized vocabularies. % like the Genomic Data Commons (GDC). 
% Finally, we outline open problems, including the need for expert-in-the-loop workflows, the variability of LLM outputs, and reproducibility concerns.
Data harmonization is an essential task that entails integrating datasets from diverse sources. Despite years of research in this area, it remains a time-consuming and challenging task due to schema mismatches, varying terminologies, and differences in data collection methodologies. This paper presents the case for agentic data harmonization as a means to both empower experts to harmonize their data and to streamline the process. We introduce \systemname, a system that combines LLM-based reasoning, an interactive user interface, and a library of data harmonization primitives to automate the synthesis of data harmonization pipelines. We demonstrate \systemname in a clinical data harmonization scenario, where it helps to interactively create reusable pipelines that map datasets to a standard format.
Finally, we discuss challenges and open problems, and suggest research directions for advancing 
our vision. 
% \vspace{-0.5em}
\end{abstract}

\maketitle

% \vspace{-1em}

% %%% do not modify the following VLDB block %%
% %%% VLDB block start %%%
% \ifdefempty{\vldbavailabilityurl}{}{
% \vspace{.15cm} %0.3cm
% \begingroup\small\noindent\raggedright\textbf{Artifact Availability:}\\
% The source code, data, and/or other artifacts have been made available at \url{\vldbavailabilityurl}.
% \endgroup
% }
% %%% VLDB block end %%%

% \vspace{-.25em}

%---------------------

% \vspace{-.10em}
\section{Introduction}
\label{sec:intro}
% \vspace{-.15em}

\begin{figure}[t]
    \centering
    \vspace{-0.75em}
    \Description{ % needed for accessibility: helps screen readers.
        The image shows an example of attributes from different data sources (T1, T2, and the GDC standard). Arrows linking the values indicate they are equivalent despite their different terminologies.
    } 
    \includegraphics[width=.8\linewidth]{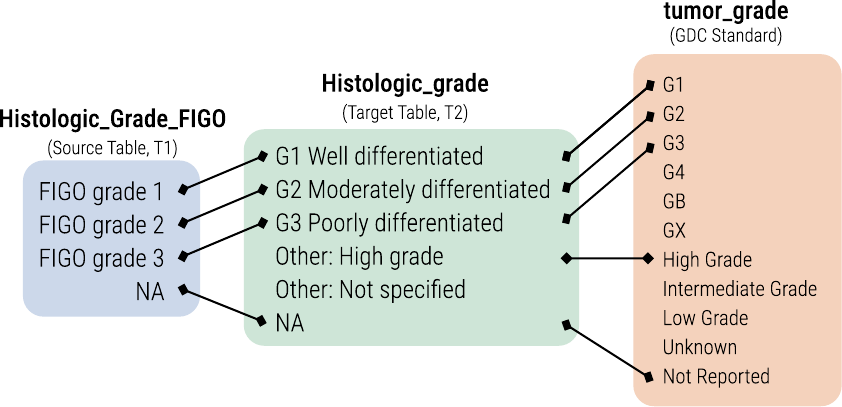}
    \vspace{-1em}
    \caption{Domain of attributes in different data sources.}
    \label{fig:tumor_grade_domains}
    \vspace{-1.5em}
\end{figure}

% Extracting insights from data remains arduous and time-consuming, often due to the challenges of integrating data from multiple sources.
Extracting insights from multiple data sources remains an arduous and time-consuming task.
% Combining data from multiple sources remains an arduous and time-consuming task.
% For example, in
In fields such as biomedicine, collecting patient data requires expensive trials that typically involve only a few dozen to hundreds of patients~\cite{li2023proteogenomic}. Since this usually leads to tables with many attributes but few samples, researchers need to combine 
% data from 
different cohorts to obtain larger sample sizes.
However, since data is often collected using differing methods,
combining them into compatible and comparable datasets often becomes a challenge~\cite{cheng2024general}. %Consider the following real examples of data harmonization in cancer research.
Consider these examples from cancer research.
% Consider the following examples.

\vspace{-.5em}
% \begin{tcolorbox}[colback=violet!2.5!white,colframe=violet!85!black]
\begin{example}[Clinical Data Harmonization] \label{example1}
% 
% T1 (mmc1.xlsx)
% Histologic_Grade_FIGO: 
% - FIGO grade 1
% - FIGO grade 2
% - FIGO grade 3
% - NA
% 
% T2 (mmc2.xlsx)
% Histologic_grade: 
% - "G1 Well differentiated"
% - "G2 Moderately differentiated"
% - "G3 Poorly differentiated"
% - "Other: High grade"
% - "Other: Not specified"
% - "NA"
% 
% GDC standard:
% G1
% G2
% G3
% G4
% GB
% GX
% High Grade
% Intermediate Grade
% Low Grade
% Unknown
% Not Reported

To obtain a larger dataset for a study on endometrial cancer, researchers aim to combine samples from two patient cohorts collected independently in two studies~\cite{dou2020proteogenomic, dou2023proteogenomic}.
% 
% These studies produced two tables containing clinical data, which we refer to as $T_1$~and~$T_2$.
% Rows in both tables represent an individual patient sample, and $T_1$ contains 179 columns while $T_2$ contains 213 columns.
These studies produced two tables containing clinical data, which we refer to as $T_1$~and~$T_2$. Each row in each table represents an individual patient sample.
Tables $T_1$ and $T_2$, contain 179 and 213 columns and 153 and 190 rows, respectively.
The goal is to combine the data to obtain a table with 392 rows.
However, even though these datasets were produced by the same research consortium, their schemas and naming standards differ significantly.
The first challenge is identifying which columns are semantically equivalent. %: there are 38,127 possible pairs in total. % 179 x 213 = 38,127
Once all pairs of equivalent columns have been identified, their values must be standardized.
Figure~\ref{fig:tumor_grade_domains} shows an example of a pair of equivalent attributes from $T_1$ and $T_2$ --  \texttt{Histologic\_grade} and \texttt{Histologic\_Grade\_FIGO}. 
% $T_1$ contains an attribute named \texttt{Histologic\_grade} with the set of unique values \{\texttt{"FIGO grade 1"}, \texttt{"FIGO grade 2"}, \texttt{"FIGO grade 3"}, \texttt{"NA"}\}
%  and $T_2$ named \texttt{Histologic\_Grade\_FIGO} contains the values 
% \{
% \texttt{"G1 Well differentiated"},
% \texttt{"G2 Moderately differentiated"},
% \texttt{"G3 Poorly differentiated"},
% \texttt{"Other: High grade"},
% \texttt{"Other: Not specified"},
% \texttt{"NA"}\}.
 % \textit{G1, G2, G3, G4}.
 The values of these attributes 
 %are not 
 are represented using different terminology.
 %the same terminology even though they are semantically equivalent.
 To produce a harmonized table, the researchers must reconcile these values 
 %into a single format 
 before merging the rows.
 For instance, they may decide that the final harmonized table $T_{target}$ will contain an attribute named \texttt{histologic\_grade} and that the format of the value used will be from \texttt{Histologic\_grade}, and thus we would need to map the values of $T_1$ to their corresponding values from $T_2$
 (e.g., 
 \mbox{\texttt{"FIGO grade 1"} $\rightarrow$\texttt{"G1 Well differentiated"}},
 \texttt{"FIGO grade 2"} $\rightarrow$ \texttt{"G2 Moderately differentiated"}, and so on).
 Note that different mappings can be applied to different attribute pairs.
% Note,~however,  that some attributes may be mapped differently.
 % that not all attributes need to be mapped like this.   E.g., attributes with unique identifiers can be kept as they appear in the original table while ensuring a uniqueness constraint.
 % \qed
\end{example}
% \end{tcolorbox}

% \begin{tcolorbox}[colback=violet!2.5!white,colframe=violet!85!black]
\vspace{-1em}
\begin{example}[Harmonizing Data to a Standard Vocabulary] \label{example2}
In a subsequent effort to foster data reuse and enable research in pan-cancer analysis~\cite{li2023proteogenomic}, researchers decided to combine 10 tables containing cohorts of a variety of cancer types \cite{cao2021proteogenomic, clark2019integrated, dou2020proteogenomic, gillette2020proteogenomic, mcdermott2020proteogenomic, huang2021proteogenomic, krug2020proteogenomic, satpathy2021proteogenomic, vasaikar2019proteogenomic, wang2021proteogenomic}.
They mapped all tables to the Genomic Data Commons (GDC) standard~\cite{gdc}, which contains attributes commonly used in cancer research.
Figure~\ref{fig:tumor_grade_domains} shows how the values for \texttt{Histologic\_grade} and \texttt{Histologic\_Grade\_FIGO} map to the corresponding GDC attribute \texttt{tumor\_grade}.
% In this case, both these attributes could be mapped to the GDC variable named \texttt{tumor\_grade} which has acceptable values 
% \{ 
% \texttt{"G1"},
% \texttt{"G2"},
% \texttt{"G3"},
% \texttt{"G4"},
% \texttt{"GB"},
% \texttt{"GX"},
% \texttt{"High Grade"},
% \texttt{"Intermediate Grade"},
% \texttt{"Low Grade"},
% \texttt{"Unknown"},
% \texttt{"Not Reported"}\}.
Once again, the acceptable values in the GDC vocabulary differ from the values used in the tables of Example~\ref{example1}.
% \qed
\end{example}
% \end{tcolorbox}

\vspace{-.5em}
\myparagraph{The Case for Agentic Data Harmonization}
Data harmonization involves several data integration tasks.
Practitioners use spreadsheet software, bespoke scripts, and significant manual work to harmonize data~\cite{cheng2024general}. These custom scripts are often not published with the data and publications, creating barriers to the reproducibility and replicability of experiments~\cite{healthcareInteroperability2025}.

%  LLM opportunities
Large language models (LLMs) open new opportunities to improve data harmonization. They can answer questions about terminology, methodologies and generate code without training data.
Recently, LLMs have shown promising results in data integration tasks, including column type annotation, schema matching, and entity linkage~\cite{magneto-vldb2025, chorus-vldb2024, feuer:vldb2024, tu2023unicorn, narayan-vldb2022}. Prompting today's frontier LLMs with a question such as ``\textit{What does FIGO grade mean?}'' reveals that they do have general knowledge about many topics, including biomedical research (the focus of our examples). This suggests that this information can be used in data harmonization tasks.

% Agentic systems
LLMs are becoming essential components for intelligent agents across various applications due to their capabilities in language understanding, tool utilization, and adapting to new information, resembling human intelligence and reasoning~\cite{yao2022react, xi2023-agents-survey, wang2024-agents-survey}. 
These capabilities have been surfaced by advancements in LLM prompting techniques that elicit reasoning, such as Chain-of-Thought~\cite{wei2022chain}
%, Tree-of-Thought~\cite{yao2024tree},
and ReAct~\cite{yao2022react}. These enable agents 
%techniques make it possible 
to handle complex tasks such as table understanding tasks through structured data manipulation~\cite{wang2024chainoftable}, and  generate reasoning traces and task-specific actions that are interleaved to complete tasks~\cite{yao2022react}.
Moreover, frameworks for building agentic systems are now widely available, such as LangChain~\cite{langchain}, Camel~\cite{li2023camel}, and AutoGen~\cite{wu2023autogen}.
%, are becoming increasingly available.

This paper presents \emph{our vision of
intelligent agents that can interact with the users and data integration algorithms to synthesize data harmonization pipelines}.  
Such agentic systems accept user task requests and prompt the user with questions required to complete a task (e.g., to request additional context or to disambiguate the input).
Similar to AutoML systems, which generate end-to-end machine learning pipelines~\cite{lopez2023alphad3m, shang2019democratizing, berti2019learn2clean}, a data harmonization agent can produce a data processing pipeline that takes as input user data and outputs a harmonized table that satisfies the user requirements. These pipelines could then be published along with research data to document the data generation process for reproducibility.

\vspace{-0.2em}
\myparagraph{Agentic Data Harmonization Challenges}
Building agentic systems presents several challenges.
First, harmonization scenarios are inherently complex, often involving difficult tasks such as schema matching, entity linkage, and data cleaning. These tasks require specialized methods to achieve high-quality results, and the scalability of these methods is critical for harmonizing large datasets.
While general-purpose LLMs offer broad capabilities, they lack transparency~\cite{pmlr-v235-huang24x} and are not optimized for these tasks, leading to inconsistent outputs, high computational costs, and performance bottlenecks, especially when handling large-scale datasets~\cite{llmDisruptVLDB2023,hsieh-etal-2023-distilling}.

Second, integrating algorithms for different tasks and generating cohesive pipelines is non-trivial.
Unlike in AutoML systems, data harmonization pipeline synthesis cannot be driven by search algorithms that optimize well-defined evaluation metrics (e.g., model accuracy) since the quality of data harmonization pipelines cannot be as easily measured using one metric.
Instead, the synthesis may need to be guided %by decisions made 
by the users, since accuracy may depend on external knowledge. For example, in the case of Fig.~\ref{fig:tumor_grade_domains}, deciding if the correct match for `\texttt{NA}' is `\texttt{Not Reported}' or `\texttt{Unknown}' may depend on the data collection methodology.
The agent should automate the laborious tasks without sacrificing accuracy and without becoming a burden: it must learn to ask questions only when necessary to avoid overwhelming the user.
To add to these challenges, LLMs are brittle: (1) they often make mistakes (also known as hallucinations) that must be identified and corrected by users; and (2) they are known to be sensitive to the prompts (i.e., small prompt changes may lead to different results)~\cite{barrie2024prompt, khattab2024dspy}.

\myparagraph{Contributions}
In this paper we discuss the opportunities and challenges for the use of agents for data harmonization. We also present  \systemname, a proof-of-concept prototype of an agentic data harmonization system that implements part of our vision. 

\systemname leverages LLM capabilities to interact with users, orchestrate specialized data integration primitives, and generate custom code when existing primitives are insufficient. These primitives implement efficient and well-established data integration algorithms that can be combined to make harmonization pipelines guided based on user feedback. Since algorithms can make mistakes, we use LLMs to evaluate the outputs of these primitives. When outputs are incorrect, the agent may take additional steps to automatically correct the errors or seek assistance from the user.
We also describe how \systemname can be applied in a real-world use case, highlighting the potential and limitations of its current design and implementation. To address these limitations, we identify open problems in this field and propose future research directions, building on previous work in machine learning and data management.

Our contributions can be summarized as follows:
% \vspace{-0.25em}
% \begin{itemize}  
%     \item We present our vision for agentic systems that help users create data harmonization pipelines by combining LLM-based reasoning, interactive user interfaces, and data integration primitives.
%     \item We build \systemname, a prototype data harmonization agent that follows our vision. It integrates \bdikit, a library of data integration algorithms, to efficiently construct harmonization pipelines. When existing functions are insufficient, \systemname leverages the LLM to dynamically generate custom code.  
%     \item We demonstrate a real-world use case where \systemname is applied to map clinical data to the GDC standard~\cite{gdc}, addressing common issues such as differing terminology and schema.
%     \item We discuss key challenges in designing data harmonization agent systems and propose a research agenda with immediate steps to address remaining open problems.  
% \end{itemize}
% 
(1) We present a vision for agentic systems that help users create data harmonization pipelines by combining LLM-based reasoning, interactive user interfaces, and data integration primitives.
(2) We introduce \systemname, a prototype data harmonization agent that  integrates \bdikit~\cite{bdi-kit-github}, a library of data integration algorithms, to efficiently construct harmonization pipelines. When existing functions are insufficient, \systemname leverages the LLM to dynamically generate custom code.  
(3) We demonstrate a real-world use case where \systemname is applied to map clinical data to the GDC standard~\cite{gdc}, addressing common issues such as differing terminology and schema.
(4) We discuss challenges in designing data harmonization agents and propose a research agenda with  steps to address open problems.

%----------------------
\section{Preliminaries}
\label{sec:preliminaries}
%----------------------

\begin{figure*}[t!]
    \centering
    \begin{subfigure}[t]{0.5\textwidth}
         \includegraphics[width=.95\linewidth]{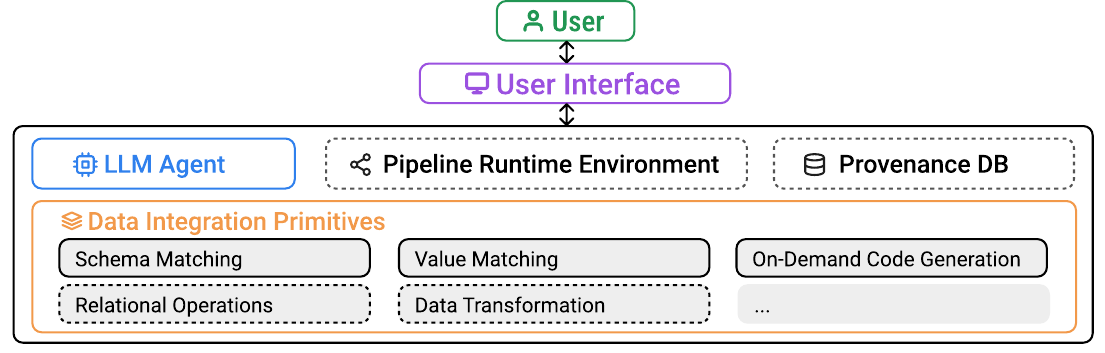}
         \hfill
         \caption{}
         \label{fig:system-diagram}
    \end{subfigure}
    \hfill
    \begin{subfigure}[t]{0.24\textwidth}
        \vbox{
            \includegraphics[width=1.0\linewidth]{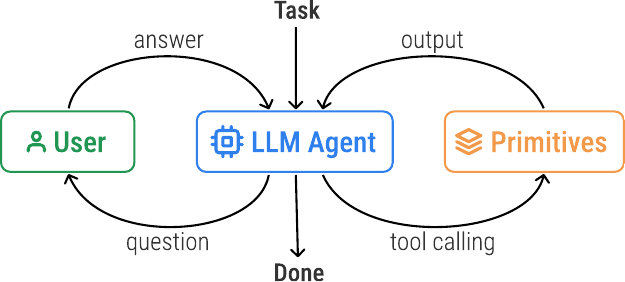}
            % \vspace{0.05in}
        }
        \caption{}
        \label{fig:agent-loop}
    \end{subfigure}
    \hfill
    \begin{subfigure}[t]{0.25\textwidth}
         % \centering
         \hfill
         \includegraphics[height=1.0in]{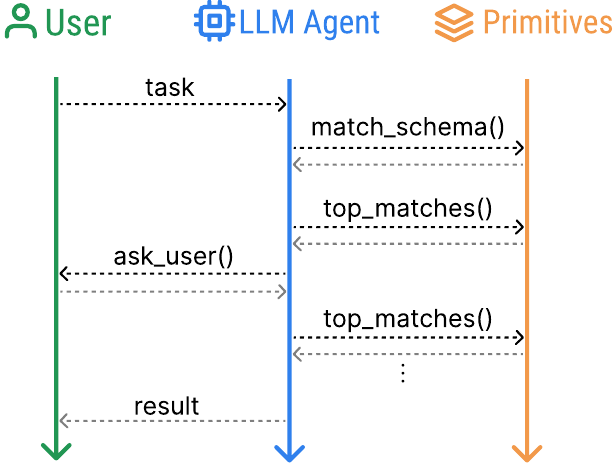}
         \caption{}
         \label{fig:sequence-diagram}
    \end{subfigure}
    \label{fig:system-roles}
    \vspace{-1em}
    \caption{
        (a) Components of an interactive agentic data harmonization system. \revised{Solid lines represent components implemented in \systemname (see Section~\ref{sec:prototype}) while dashed lines represent components not yet implemented.}
        (b) The LLM agent loop: given a task, the LLM will repeatedly execute actions (call tools or ask questions to the user) until the task is completed.
        (c) A sequence diagram illustrating an example of the communication workflow between the user, agent, and primitives.
    }
    \vspace{-.4cm}
\end{figure*}

Broadly speaking, data harmonization refers to the practice of combining and reconciling different datasets to maximize their comparability or compatibility~\cite{cheng2024general}.
Although data harmonization goals can vary greatly depending on the data modalities and goals involved, this paper focuses on the following problem:

\vspace{-.3cm}
\begin{definition}[Tabular Data Harmonization]
Given a set of source tables $T_{1}$, $T_{2}$, ..., $T_{n}$ and a target schema $S_{target}$ composed of a set of attributes, each attribute specifying its acceptable values, i.e., a domain, the goal is to derive a computational pipeline $\mathcal{P}$ that takes as input the source tables and applies transformation functions to values and combines tables (e.g., using union and join operations) to generate an output table $T_{target} = \mathcal{P}(T_1, T_2, ..., T_n)$ that adheres to the given target schema $S_{target}$.
\end{definition}
\vspace{-.3cm}

Note that instead of aiming at maximizing an objective measure of comparability or compatibility between input and target, we assume the existence of the canonical data representation $S_{target}$. 
As seen in Examples~\ref{example1} and ~\ref{example2}, this specification $S_{target}$ consists of a set of attribute names and domain specifications that can either come from a standard data vocabulary (such as the GDC) or could be derived from the existing data in the source tables.

Data harmonization pipelines can be of multiple forms. A simple example is a linear sequence of data processing operations (e.g., as in scikit-learn pipelines~\cite{sklearn-pipelines} or Jupyter Notebooks~\cite{jupyter}). However, they can also be declaratively represented as direct acyclic graphs (DAG), such as in AutoML systems~\cite{lopez2023alphad3m, shang2019democratizing}.
In practice, data harmonization pipelines are typically complex, custom scripts created iteratively through a trial-and-error process~\cite{cheng2024general}.
It is challenging to code and manually explore the space of pipelines to identify the most effective one for a given task. For subject matter experts without programming experience, creating these pipelines is simply out of reach. We envision that LLM-based agents can help address these challenges as we elaborate next.

\vspace{-0.75em}
%-------------------------------------
\section{Agentic Data Harmonization}
\label{sec:agentic-data-harmonization}
%-------------------------------------

% \input{text/figures}

% LLM-based agents integrate the capabilities of language models with reasoning, planning, memory retention, and interaction with external tools~\cite{qin2024toolllm,code-generatingLLMCHI23}.
We propose using LLM-based agents to facilitate the interactive construction of harmonization pipelines through natural language and visual interfaces. We aim to simplify the harmonization process and empower domain experts to harmonize their data~effectively.

Our approach has three main components: \textit{harmonization primitives}, \textit{harmonization agents}, and \textit{human-agent interaction} (Figure~\ref{fig:system-diagram}). 
The system supports a two-way interaction between \textit{users} and \textit{agent}. While users drive the system and define the tasks to be performed, the agent aims to automate the harmonization tasks, leaving to users only decisions that need external context.
By recording user-agent interactions and the derived computational pipelines, we maintain provenance of the harmonization process, supporting transparency and reproducibility and making it possible to publish the harmonized data with the pipeline used to derive it.

\myparagraph{Harmonization Primitives}
The bottom of Figure~\ref{fig:system-diagram} shows a library of components that we refer to as \textit{data integration primitives}. These are algorithms or routines that solve well-defined data integration tasks such as schema matching and value mapping. % function-attribute matching,
Some of these are lower-level routines that support other higher-level tasks, e.g., column-type annotation may be a component of schema-matching algorithms~\cite{feuer:vldb2024}, while entity resolution and deduplication may be key to performing data standardization~\cite{christophides2020overview}. The set of primitives can be heterogeneous and evolve to support the system capabilities.
An important requirement is that they need to be \textit{composable} and invoked by both user and AI agents.

\revised{

% \textit{Primitive composability} refers to the ability to combine primitives to create a pipeline: the output of a primitive $p_1$ can be used as input of another primitive $p_2$. For example, \textit{schema matching} primitives can output a list of source-target attribute pairs.
% These lists form the input for the \textit{value matching} primitives, which then find equivalences between the values of the source attribute to the target. Finally, the output of value-matching primitives can be used to assemble a \textit{harmonization specification} that describes the transformation of source tables $T_i$ into a target output table $T_{target}$.
% In short, primitive composability enables the creation of data harmonization pipelines by allowing the chaining of operations that take source tables as input and produce harmonized data as output.
\textit{Composability} is the ability to combine different primitives to create a data harmonization pipeline. For instance, output from schema matching primitives, which generates source-target attribute pairs, can feed into value mapping primitives that find equivalences between the values of those source and target attributes. Finally, the output of value mapping primitives can be used to create a \textit{harmonization specification} that describes the transformation of source tables $T_i$ into a target output table $T_{target}$.
}

\myparagraph{Data Harmonization Agents}
% An agent is an entity that senses its environment and acts on it~\cite{russell2016artificial}. In our context, we define a \textit{data harmonization agent} as a program that interacts with its environment (e.g., data, primitives, users) to gather information and autonomously performs tasks to complete data integration tasks.
% Similar to autonomous driving systems, data harmonization agents can operate at varying levels of autonomy~\cite{society2018taxonomy}, ranging from systems that offer no assistance—leaving all tasks to the user—to fully autonomous agents that handle the entire harmonization process independently.
% In Section~\ref{sec:prototype}, we focus on interactive agents that react to a users' query, but more autonomy is possible in our framework.
An agent senses its environment and acts upon it~\cite{russell2016artificial}. In our context, we define a \textit{data harmonization agent} as a program that interacts with its environment (e.g., data, primitives, users) to autonomously complete harmonization tasks. 
%As in autonomous driving systems, 
Agents can operate at various levels of autonomy~\cite{society2018taxonomy}, from offering minimal user assistance to fully independent harmonization. In Section~\ref{sec:prototype}, we focus on interactive agents that respond to user queries, though more autonomy is possible within our framework.

\revised{
% To fulfill user requests, the agent must first decompose the problem into a series of actions. These actions can be of various types, including the execution of existing integration primitives (tool calling) or the execution of code generated on demand.
To fulfill user requests, the agent must first decompose the problem into a series of actions of various types, including (1) the execution of existing integration primitives (\textit{tool calling}) or (2) code generated on demand.}
To decompose the problem and decide on what action to take, the harmonization agent leverages an LLM, which is given descriptions of the task and integration primitives available for use. 
The LLM then repeatedly returns actions (e.g., a code snippet) that can be executed in a runtime environment (e.g., a Python kernel). Action outputs are fed back into the LLM, which decides if additional actions are needed.
As illustrated in Figure~\ref{fig:agent-loop}, this loop executes until the task is deemed complete.

The main loop is orchestrated by a driver code that takes inputs from the user (i.e., prompts) that describe the task to be performed. 
This driver is also responsible for (1) communicating with users to request inputs (e.g., when the LLM asks for task clarification or user preferences) and (2) managing the state (memory) of the agent. For example, it can also track and store the history of actions and user interactions in a \textit{Provenance DB}.
This data can support decisions about future actions, and transparency, and be used to generate harmonization \textit{specifications} or scripts to reproduce the results.

Given the complexity of harmonization tasks, it is crucial to have high-level primitives available as building blocks for the pipeline.
This allows encoding prior knowledge and using efficient algorithms known to be effective for a specific task.
These primitives encompass algorithms for tasks such as schema matching, entity resolution, and value mapping. Of course, primitives can go beyond hard-coded functions that implement deterministic algorithms. 
For instance, they can be workflows that use LLMs to perform specific tasks such as in \cite{magneto-vldb2025} or they could generate code on demand (e.g., to extract data from or to transform attribute values~\cite{autoformula2024}).

\myparagraph{Human-Agent Interaction}
A key aspect of our architecture is the interaction between users and agents. 
In Section~\ref{sec:prototype}, we show that our current prototype is proficient in text-based conversational interactions: it can parse users' input and act on them.
However, to make systems more effective, we argue that future systems need to support graphical user interfaces that help users (1) to reason about the input data and the agent's output, and (2) to refine the task definition and pipelines derived by the agent.
% In this paper, we focus on interactions based on natural language via a chatbot interface. However, it should be possible to implement even more intuitive and efficient graphical interfaces that combine natural language with point-and-click components. 
For example, as done in interactive AutoML tools~\cite{santos2019visus}, the system could guide the user through the harmonization process, recommend the available actions, track progress, and provide data visualizations to help the user better make sense of the data. This may help prevent common issues in natural language such as ambiguity~\cite{esfandiarpoor2024followup,yu2019flowsense}.

% We argue that systems must go beyond text-based conversational interactions: they need to support rich visual data representations and should allow interactions that help users reason about the answers produced by the agent and refine the task definition as well as the pipelines derived by the agent. 
% In complex tasks such as harmonization, user interaction is necessary since many decisions needed to complete the task (e.g., whether two terms represent the same concept) can be difficult or impossible even for domain experts. For instance, it may require knowledge not available in data, such as the context in which data was collected or the purpose of the harmonization.
In complex tasks like harmonization, user interaction is crucial because many decisions---such as whether two terms represent the same concept---can be difficult even for domain experts. These judgments often depend on context not captured in the data, such as how it was collected or the purpose of the harmonization.
Furthermore, as in any automated data integration system, LLM-based agents can make mistakes. 
% Together with user interfaces, users can provide guardrails and safety mechanisms to prevent downstream problems.
When combined with well-designed user interfaces, users are better equipped to guide the agent, providing guardrails and safety mechanisms that neither users nor interfaces could achieve as effectively on their own. %-- leading to more reliable and accurate outcomes.

\vspace{-.75em}
\section{System Prototype \& Use Case} \label{sec:prototype}

% \revised{We implemented a proof-of-concept for the harmonization agent shown in Figure~\ref{fig:system-diagram}. Next, we provide implementation details and illustrate how the system works through a concrete example.}
\revised{We implemented a proof-of-concept of our vision (Figure~\ref{fig:system-diagram}). In what follows, we illustrate how it works with a concrete example.}

\myparagraph{Data Integration Primitives}
We used data harmonization primitives from \bdikit \cite{bdi-kit-github}, an open-source Python library that we designed with the explicit goal of composability.
Currently, it includes implementations of multiple \textit{schema matching} and \textit{value mapping} algorithms using a composable API. It also includes several classic algorithms from Koutras et al.~\cite{koutras2021valentine} and language model-based algorithms~\cite{liu2024enhancing,magneto-vldb2025}. Most functions take as input a \texttt{source} parameter that represents the user's input DataFrame and returns the output formatted as another DataFrame. The \texttt{target} parameter can either be a string representing a target standard schema (e.g., \texttt{`gdc'}) or a target DataFrame, this allows switching between the two tasks described in Examples~\ref{example1} and ~\ref{example2}. 
\revised{Figure~\ref{fig:bdi-kit} shows a sample of  functions integrated in \systemname. The source code is on GitHub~\cite{harmonia-github}.}

\begin{figure}[b]
\vspace{-2.2em}
\centering
\begin{tcolorbox}[colback=black!2.5!white,colframe=black!85!black,boxrule=0.25mm,boxsep=4pt,left=0pt,right=0pt,top=0pt,bottom=0pt]

    \footnotesize
    
    \texttt{\textbf{match\_schema}(source, target, method, ...)} \\
    Maps the schema of a source table to a target schema (table or predefined standard like \texttt{gdc}) using a specified method.
    \vspace{.5em}

    \texttt{\textbf{top\_matches}(source, column, target, top\_k, method, ...)} \\
    Finds the top-\texttt{k} matches between a source column and columns of a target schema.
    \vspace{.5em}

    \texttt{\textbf{match\_values}(source, target, column\_mapping, method, ...)} \\
    Matches values between columns of a source and a target using a specified method, returning one or more result tables.
    \vspace{.5em}

    % \texttt{\textbf{top\_value\_matches}(source, target, column\_mapping, top\_k,...)} \\ %  method, 
    % Finds the top-\texttt{k} value matches between columns in a source and target.
    % \vspace{.5em}

    % \texttt{\textbf{view\_value\_matches}(matches, edit = False)} \\
    % Displays value match results in a table format, with optional editing.
    % \vspace{.5em}

    % \texttt{\textbf{preview\_domain}(dataset, column, limit = None) → DataFrame} \\
    % Previews unique values and descriptions in a specified column of a dataset.
    % \vspace{.5em}

    % \texttt{\textbf{merge\_mappings}(mappings, user\_mappings = None) → List} \\
    % Combines computed and user-provided mappings into a plan for data transformation.
    % \vspace{.5em}

    \texttt{\textbf{materialize\_mapping}(input\_table, mapping\_spec) → DataFrame} \\
    Transforms a source table into a new table using a mapping specification.
    % 
    % \texttt{create\_mapper(input) → ValueMapper} \\
    % Creates a mapper object for transforming column values based on the input type. \\
    
    % \texttt{MappingSpecLike} \\
    % Defines mappings between source and target columns, including optional value transformations. \\
\end{tcolorbox}
\vspace{-1.5em}
\caption{Some \bdikit functions integrated in \systemname.}
\vspace{-0.5em}
\label{fig:bdi-kit}
\end{figure}

% JF: trying to make parag title consistent with figure
\myparagraph{LLM Agent \& User Interface} 
%We implemented \systemname, a system prototype that uses the data integration primitives from Figure~\ref{fig:bdi-kit} and interacts with users via a text-based chat box. 
Implementing an agent entails writing carefully crafted function and task descriptions that are combined to assemble system prompts fed to the LLMs. For our prototype, we implemented tool wrappers for each of the \bdikit functions, along with descriptions of when each should be used. We also provide descriptions of the data harmonization steps, when the LLM should request help from the user, and output formatting instructions. \revised{
These descriptions are available in the code~\cite{harmonia-github}.}

\revised{
To implement tool calling, we used the Archytas~\cite{archytas}, an open-source library for building AI agents based on the ReAct framework~\cite{yao2022react}, and used the \texttt{GPT-4o} model (i.e., no fine-tuning is involved). Instead of using a fixed pipeline that invokes the LLM, we implement a dynamic agentic loop, as described in Figure~\ref{fig:agent-loop}. The LLM has the flexibility to break down the task into multiple tool call instructions needed to complete the task.
% , a contextually-aware notebook system that supports the development of chatbot user interfaces. 
% This allows us to implement features that facilitate interaction, such as rendering markdown-formatted outputs in the browser, displaying the reasoning steps taken by the agent, collecting user inputs, and feeding them back to the agent.
The user interface is implemented using Beaker~\cite{beaker} and is currently limited to a chat-based interface and workflow that resembles a Jupyter notebook. The system renders markdown-formatted outputs in the browser, displays the reasoning steps taken by the agent, and collects user inputs using text input boxes and feeds them back to the agent. A video demonstrating these interactions is available online \cite{harmonia-video}.}

%\myparagraph{Use case: harmonization with the GDC standard} % previous title
\myparagraph{Use Case: Harmonizing a Dataset with the GDC Schema}
To demonstrate the ability to produce good data harmonization plans, we present a use case that harmonizes a dataset from Dou et al.~\cite{dou2020proteogenomic} (described in Example~\ref{example1}) with the GDC standard. 
% Below we provide an example of interaction with the chatbot agent.
Below, we show message exchanges between the user ({\scriptsize \faUser}) and the agent ({\scriptsize \faRobot}), along with explanations of the actions taken by the agent to answer user queries. \revised{As the interactions show, \systemname chooses proper tools to solve the task, which includes generating and executing Python code, as well as \bdikit functions with the appropriate parameters.}

The user starts by requesting the system to load a CSV file that contains the attributes of interest:
% \vbox{
\vspace{-.5em}
\chatboxUser{
Load the file dou.csv as a dataframe and subset it to the following columns: Country, Histologic\_Grade\_FIGO, Histologic\_type, FIGO\_stage, BMI, Age, Race, Ethnicity, Gender, Tumor\_Focality, Tumor\_Size\_cm.
}
\vspace{-.5em}
% \vspace{-1em}
% \chatboxAgent{
% The file 'dou.csv' has been successfully loaded and subsetted to the specified columns: Country, Histologic\_Grade\_FIGO, Histologic\_type, FIGO\_stage, BMI, Age, Race, Ethnicity, Gender, Tumor\_Focality, Tumor\_Size\_cm.\\
% Outputs:\\
% {\color{gray}[...data frame displayed here...]}
% }
% \vspace{-.5em}
% }
\noindent \revised{As a response, the agent shows a success message and a table preview. To produce this result, 
% Responding to this prompt,
} the agent automatically (1) generates Python code using the Pandas library that loads and selects the requested columns and (2) executes the generated code to load the CSV file into memory. This example illustrates the capability of \textit{on-demand code generation} to accomplish a requested task.

Next, the user requests the system to perform schema matching between the table loaded and the attributes from the GDC standard:

\vspace{-.5em}
\chatboxUser{
\revised{
Match this to the GDC schema and fix any results that don't look correct.}
}
\vspace{-.5em}

\noindent
As shown in Figure~\ref{fig:sequence-diagram}, the system interacts with an LLM multiple times to answer this request. First, it sends the initial user's prompt to the LLM, which recognizes that the function \texttt{match\_schema} is adequate to solve the request, and thus returns a tool calling instruction containing the \texttt{match\_schema} function name along with the needed parameters.
After receiving the instruction, the system executes the function, collects matches, and returns them to the LLM. 
Note that the LLM interprets the user's query and correctly identifies \revised{the function arguments: it uses \texttt{gdc} as the target schema and a default value for the \texttt{method} parameter in \texttt{match\_schema()}. 
Alternatively, the agent also supports requesting a specific method in the prompt, offering experts greater control and customization.s
}

After analyzing the column match results received in the follow-up prompt, the LLM detects a mapping mistake for the column \texttt{Histologic\_type}, which is incorrectly mapped to \texttt{roots}. To fix the error, the LLM triggers a tool calling instruction that uses the primitive function \texttt{top\_matches} to find the top-10 alternative candidates for \texttt{Histologic\_type}. After analyzing the results, it suggests using \texttt{primary\_diagnosis} as a correction to the user. % \revised{(who replies ``yes''):}
% (who agrees) as seen in the following messages:

\vbox{
%\vspace{-.5em}
\chatboxAgent{The current mapping for `Histologic\_type' is `roots', which seems incorrect. The top alternatives are:\\
- history\_of\_tumor\_type\\
- primary\_diagnosis\\
- sample\_type\\
I suggest using `primary\_diagnosis'. Do you agree with this choice?
}
\vspace{-.25em}
% \chatboxUser{yes}
}
\vspace{-.5em}
\noindent 
\revised{This exemplifies an important benefit of the agent in augmenting and improving the effectiveness of the underlying libraries: using its knowledge, it is able to detect the error; then, it leverages the library to identify alternative choices. This also shows an example of complex task planning enabled by the agentic loop (Figure~\ref{fig:agent-loop}).}

%As a final response to the task, 
The agent then displays a table of column mappings, where the ``\texttt{Source Column}'' contains columns from the source table (\texttt{dou.csv}), ``\texttt{Target Column}'' contains a column match from the GDC standard, and ``\texttt{Corrected?}'' indicates whether the LLM has corrected the original \bdikit match.

\vspace{-.25em}
\chatboxAgent{
Result:\\

\vspace{-1em}
\hspace{-0.25cm}
\begin{tabular}{lll}
\textbf{Source Column} & \textbf{Target Column} & \textbf{Corrected?} \\
{\color{gray}[...]} & {\color{gray}[...]} & {\color{gray}[...]} \\
Country & country\_of\_birth & No \\
Histologic\_Grade\_FIGO & tumor\_grade & No \\
Histologic\_type & primary\_diagnosis & Yes, from ``roots'' \\
Tumor\_Size\_cm & tumor\_depth\_measurement & No \\
\end{tabular}
}
% }
\vspace{-.5em}
Since the user is a domain expert familiar with the GDC standard, they ask for
%and knows that there may be 
better matching options for \texttt{Tumor\_Size\_cm}.
%, it asks the system to find an alternative match. 
The system updates the previous list with a more appropriate match found using the \texttt{top\_matches} function:
%. This interaction is shown next:

\vbox{
\vspace{-.25em}
\chatboxUser{
Find alternative mappings for Tumor\_Size\_cm.
}
\vspace{-1em}
\chatboxAgent{Result:\\

\scriptsize
\vspace{-1em}
\hspace{-0.25cm}
\begin{tabular}{lll}
    \textbf{Source Column} & \textbf{Target Column} & \textbf{Corrected?} \\ 
    {\color{gray}[...]} & {\color{gray}[...]} & {\color{gray}[...]} \\
    % Country & country\_of\_birth & No \\ 
    % Histologic\_Grade\_FIGO & tumor\_grade & No \\ 
    % Histologic\_type & primary\_diagnosis & Yes, from "roots" \\ 
    % FIGO\_stage & figo\_stage & No \\ 
    % BMI & bmi & No \\ 
    % Age & age\_at\_onset & No \\ 
    % Race & race & No \\ 
    % Ethnicity & ethnicity & No \\ 
    % Gender & gender & No \\ 
    Tumor\_Focality & tumor\_focality & No \\ 
    % Tumor\_Size\_cm & tumor\_largest\_dimension\_diameter & Yes, from "tumor\_depth\_measurement" \\ 
    Tumor\_Size\_cm & tumor\_largest\_dimension\_diameter & Yes, from "tumor\_depth... \\ 
\end{tabular}
}
} %\vbox
\vspace{-0.5em}
% --------------
% Value mappings
% --------------

\noindent After finding correct matches for all source columns, the next step is to find value mappings for each column match. The following prompt requests the agent to find such mappings:

\vspace{-0.5em}
% \vbox{
\chatboxUser{\raggedright Find value mappings for the columns Country, Histologic\_Grade\_FIGO, Histologic\_type, FIGO\_stage, Race, Ethnicity, Gender, Tumor\_Focality. If there are any errors in the mappings, please provide suggestions.
}
\vspace{-0.5em}
\noindent
As in the previous prompts, finding value mappings requires multiple LLM interactions. First, the LLM call the \texttt{match\_values()} function for each of the source-target column pairs using the default method \texttt{tfidf}  (an efficient algorithm that chooses matches based on the tf-idf similarity over n-grams). After receiving the value mappings, the LLM verifies if they are correct and fixes the erroneous entries. To do so, it calls another function to discover the list of acceptable values in a GDC attribute and picks the correct one. A snippet of the result is shown next (we omit part of the output due to lack of space):
%in the box below:

\vspace{-0.5em}
\chatboxAgent{\scriptsize
{\color{gray}[...]}\\

\vspace{-0.5em}
\textbf{Histologic\_Grade\_FIGO → tumor\_grade}\\

\vspace{-0.5em}
\hspace{-0.25cm}
\begin{tabular}{lll}
    \textbf{Source Value} & \textbf{Target Value} & \textbf{Corrected?} \\ 
    FIGO grade 1 & G1 & Yes, from "Low Grade" \\ 
    FIGO grade 2 & G2 & Yes, from "Low Grade" \\ 
    FIGO grade 3 & G3 & Yes, from "Low Grade" \\ 
\end{tabular}\\

{\color{gray}[...]}\\

\vspace{-0.5em}
\textbf{FIGO\_stage → figo\_stage}\\

\vspace{-0.5em}
\hspace{-0.25cm}
\begin{tabular}{lll}
    \textbf{Source Value} & \textbf{Target Value} & \textbf{Corrected?} \\ 
    IIIB & Stage IIIB & No \\ 
    IIIA & Stage IIIA & No \\ 
    II & Stage II & Yes, from "Stage III" \\ 
\end{tabular}\\

\vspace{-.25em}
%{\color{gray}[...]}
}
% }

\noindent
Note that the LLM detects and corrects errors automatically without user intervention. For example, it corrects the match $\langle\texttt{II}, \texttt{Stage III}\rangle$ to $\langle\texttt{II}, \texttt{Stage II}\rangle$. 
This shows how we can leverage efficient algorithms to perform most of the work while resorting to LLMs as evaluators that can identify and fix mistakes only when needed.

% ------------------
% Materialize schema
% ------------------

The final step is to generate a harmonized table:
%. The boxes below show the user prompt as well as the agent response:

\vbox{
\vspace{-0.25em}
\chatboxUser{Please create a final harmonized table based on the discovered column and value mappings and save it at "dou\_harmonized.csv".}
\vspace{-1em}
\chatboxAgent{The harmonized table has been successfully created and saved as `dou\_harmonized.csv'.}
\vspace{-0.25em}
}
\noindent
Here, the agent uses \texttt{materialize\_mapping} to generate a table and save file \texttt{`dou\_harmonized.csv'}. The LLM  first creates a declarative mapping specification according to the \bdikit's format used to describe a harmonization plan. Figure~\ref{fig:mapping-spec} shows a snippet of the generated harmonization plan, which includes a declarative specification of the mappings in JSON. In this example, we only show dictionary-based transformations that map source values into target values. However, \bdikit also supports other types of transformations, such as custom mappings that take as input a custom Python function (or lambda).
This can potentially be used along with the on-demand code generation.

The main advantage of using a declarative language to describe harmonization plans is that it enables reproducibility: once a plan is created, users can feed the plan along with the source data into \texttt{materialize\_mapping} function to recreate the harmonized data. This does not require re-running any LLM-based interactions, since all transformations are encoded in the harmonization plan.

% % This is a hack to remove the syntax highlighting error in the minted box below
% \AtBeginEnvironment{minted}{\renewcommand{\fcolorbox}[4][]{#4}}
% \begin{figure}[h]
% \vspace{-.75em}
% \begin{tcolorbox}[colback=black!2.5!white,colframe=black!85!black,boxrule=0.25mm,boxsep=4pt,left=0pt,right=0pt,top=0pt,bottom=0pt]
% \begin{minted}[fontsize=\scriptsize]{json}
% [ 
%  [...]
%  { "source": "Country",
%    "target": "country_of_birth",
%    "matches": [ ["United States", "United States"],
%                 ["Poland", "Poland"],
%                 ["Ukraine", "Ukraine"] ] },
%  { "source": "Histologic_Grade_FIGO",
%    "target": "tumor_grade",
%    "matches": [ ["FIGO grade 1", "G1"],
%                 ["FIGO grade 2", "G2"],
%                 ["FIGO grade 3", "G3"] ] },
%  [...]
% ]
% \end{minted}
% \end{tcolorbox}
% \vspace{-1.25em}
% \caption{A snippet from the mapping specification generated that is passed to \texttt{materialize\_mapping()} function.}
% \label{fig:mapping-spec}
% \end{figure}

% This is a hack to remove the syntax highlighting error in the minted box below
\AtBeginEnvironment{minted}{\renewcommand{\fcolorbox}[4][]{#4}}
\begin{figure}[h]
\vspace{-.75em}
\begin{tcolorbox}[colback=black!2.5!white,colframe=black!85!black,boxrule=0.25mm,boxsep=4pt,left=0pt,right=0pt,top=0pt,bottom=0pt]
\begin{minted}[fontsize=\scriptsize]{json}
[ 
 [...]
 { "source": "Histologic_Grade_FIGO",
   "target": "tumor_grade",
   "matches": [ ["FIGO grade 1", "G1"],
                ["FIGO grade 2", "G2"],
                ["FIGO grade 3", "G3"] ] },
 [...]
]
\end{minted}
\end{tcolorbox}
\vspace{-1.25em}
\caption{A snippet from the mapping specification generated that is passed to \texttt{materialize\_mapping()} function.}
\vspace{-1.0em}
\label{fig:mapping-spec}
\end{figure}

\revised{
\myparagraph{Evaluation} We conducted a preliminary evaluation and compared \systemname against baseline methods from \bdikit executed without agent support.
%of the schema matching and value mapping tasks using the attributes from Dou~et~al.~\cite{dou2020proteogenomic}. \systemname was compared against baseline methods (the \bdikit functions) executed without the agent's support.
The results, summarized in Table~\ref{tab:performance_comparison}, show that \systemname achieved the best performance across both tasks. For schema matching, \systemname successfully 
%corrected and 
associated the column \texttt{`Histologic\_type'} with \texttt{`primary\_diagnosis'}, which contains 2,625 unique values in the GDC. For value mapping, \systemname was able to correct values such as \texttt{`FIGO grade 1'} to \texttt{`G1'}.

\begin{table}[b]
    \vspace{-1.5em}
    \centering
    \small
    \caption{Performance comparison for schema matching and value mapping tasks.}
    \vspace{-1em}
    \begin{tabular}{l|cccccc}
        \toprule
        \textbf{Task} & \textbf{Method} & \textbf{Accuracy} & \textbf{Precision} & \textbf{Recall} & \textbf{F1} \\
        \midrule
        %\multirow{2}{*}{Schema Matching} 
        Schema & Baseline & 0.88 & 0.78 & 0.78 & 0.78 \\
        Matching& Harmonia & \textbf{1.00} & \textbf{1.00} & \textbf{1.00} & \textbf{1.00} \\
        \midrule
        %\multirow{2}{*}{Value Mapping} 
        Value& Baseline & 0.58 & 0.58 & 0.59 & 0.57 \\
        Mapping & Harmonia & \textbf{0.68} & \textbf{0.69} & \textbf{0.69} & \textbf{0.68} \\
        \bottomrule
    \end{tabular}
    \label{tab:performance_comparison}
    \vspace{-2em}
\end{table}

}

\section{Research Opportunities}
\label{sec:research-agenda}

\systemname shows the potential of LLM-based agents to orchestrate actions, evaluate function outputs, detect errors, and generate additional functions. However, there are still open research questions to expand the system's capabilities and improve its effectiveness and usability, which we outline below.
%Below we outline some of the immediate steps towards these opportunities.

% ----------------------------------------------------
\myparagraph{Agent Evaluation \& Benchmarks} 
Most existing evaluation benchmarks are focused on isolated tasks, such as schema matching or entity linking~\cite{koutras2021valentine, magneto-vldb2025, wang2021machamp}. 
However, agentic systems create a need for end-to-end evaluation benchmarks and metrics to measure progress effectively. Recently, researchers have started developing benchmarks for evaluating agents in various tasks, including data analysis and ML engineering~\cite{chan2024mle, hu2024infiagent, zhang2024benchmarking, huang2024mlagentbench}. 
To spur and measure the progress in data harmonization, we need to create benchmarks tailored for this task.

\myparagraph{Data Integration Primitives}
\label{sec:proposed-primitives}
 \emph{Uncertainty} quantification and \emph{explanations} should be exposed by the primitives to guide decision-making~\cite{uncertainty2009}.
Uncertainty in data integration arises from factors like ambiguous schema mappings and data values~\cite{WangHM2018}. \systemname tackles this by exposing similarity scores through its primitives. For example, a value matcher can return similarity scores so that the agent can trigger complementary primitives (e.g., value mapping) for deeper analysis. A key challenge is conveying the meaning of uncertainty measures from diverse primitives to LLMs and end users and instructing LLMs on how to use them~\cite{dagstuhl2029explanation}.

Also, LLMs often lack transparency~\cite{pmlr-v235-huang24x}, so primitives must provide \textit{interpretable explanations} to promote user trust. Primitives should offer clear usage documentation and expose their decision rationale. For instance, a matching algorithm description could document whether its similarity scores derive from syntactic similarity, semantic embeddings, or value distribution analysis. LLMs can also explain their decisions based on domain knowledge and primitive instructions, helping users better understand why a particular path was chosen~\cite{explainability2024,TableMeetsLLM2024}. A key opportunity is training agents to discern when to rely on LLM explanations, apply alternative strategies, or engage with the user directly.

Mapping data between schemata involves resolving entities and transforming data~\cite{dataexchange2018}. LLM-based methods have proven helpful in entity resolution~\cite{narayan-vldb2022, fan2024cost}, generating or finding transformations functions~\cite{zdnet-github-copilot, trummer2022codexdb,autotables2023, autoformula2024, dtt2024, SheetAgent2024}, and evaluating LLM performance on such tasks~\cite{ma2024spreadsheetbench}.
However, integrating these methods into agent-based systems requires consistency across diverse data models and alignment with broader agent goals.
Also, we need methods that allow agents to identify and recommend appropriate attribute transformations and suitable functions for a given input dataset. This is especially in challenging cases such as table restructuring or non-standard formats~\cite{autotables2023}.
%Another immediate step is to curate a library of transformation functions specialized in data transformation tasks that agentic systems can readily reuse.

\myparagraph{Robustness and Reliability}
LLMs have shown inconsistency across various scenarios (e.g., as text summarization evaluators~\cite{stureborg2024large}), often producing varying results when executed multiple times~\cite{barrie2024prompt}. This variability can undermine reproducibility and reliability, particularly in critical applications where consistent mappings or transformations are crucial. In our experiments, we observed that while the LLM typically identified and fixed incorrect mappings, it occasionally failed to do so (even when provided with the same prompts).

Handling large and complex tables with many attributes poses additional challenges, as these can lead to long chat histories that exceed the LLM context window. When this occurs, the LLM may lose access to earlier relevant information, thereby affecting the robustness. While approaches have been proposed to mitigate context window limitations (e.g., \cite{jin2024llm} \cite{ma2024megalodon}), 
it is not clear if they address the issues in agent systems. 
Equipping agents with access to read and store data in external databases (such as the Provenance DB discussed in Section ~\ref{sec:agentic-data-harmonization}) may be an effective solution to this issue.

\myparagraph{User-Agent Interaction and Interfaces} To improve usability, agentic systems must go beyond natural language (NL) interfaces. While NL is flexible, it is also often ambiguous and may lead to under-specified task descriptions~\cite{zhang2023clarify}. 
Since the same task can be expressed in multiple unpredictable ways, a mismatch between the user task descriptions and agent prompt specifications may occur. Therefore, detecting when clarifications are needed may help increase overall success~\cite{zhang2023clarify}.
These issues could also be potentially addressed by action-oriented UIs that recommend actions linked to predefined prompts. Moreover, using rich visual representations may be more effective at conveying information to the user.

\myparagraph{Provenance-Aware Agents} 
Provenance-enabled systems have demonstrated promising results in data science pipelines ~\cite{rupprechtVLDB2020,chapmanCapturing2020}.
In data harmonization pipelines, we can track all interactions that contribute to obtaining a specific value mapping. For example, we could record all user-agent and agent-primitive interactions involved in determining the mapping of ``\texttt{FIGO grade 1}'' to ``\texttt{G1}'' (see Figure \ref{fig:mapping-spec}). 
This would allow tracing the lineage of all values in the output data. Moreover, this information could potentially be used to learn user preferences that reduce the need for user interactions~\cite{koop2008viscomplete}.
% Such systems not only enhance transparency but also reduce the need for user interactions, particularly for new practitioners. 
By learning from provenance and pipelines accumulated over time, a system could further streamline the harmonization process by automating all steps and presenting final results directly to users.
% For instance, a provenance system could streamline the harmonization process by bypassing intermediate steps and presenting final results directly to new users.

\myparagraph{Data Harmonization Pipelines} 
In our system, data harmonization is expressed as a pipeline where multiple primitives (predefined, user-defined, or agent-defined) are interconnected through their inputs and outputs to produce the final harmonized dataset. 
% The pipeline can have various objectives, such as creating a dataset with a desired minimum number of rows, returning a sample result in a few seconds, or generating a high-quality dataset based on a chosen quality metric. 
The pipeline can have various objectives, such as maximining the number of correct column matches and value matches, minimizing the number of interactions with users, or minimizing the computational costs (e.g., runtime or LLM calls).
Achieving these objectives represents an optimization problem, requiring the system to navigate a complex search space and balance multiple objectives to determine an optimal sequence of operations.
%Recent work has explored large search spaces for building end-to-end pipelines for various tasks~\cite{lopez2023alphad3m,substrat2022,autopipeline2012vldb,volcanoVLDB2021}. 
This is similar to the process used by AutoML systems that automatically synthesize end-to-end pipelines~\cite{lopez2023alphad3m}.
%with the goal of constructing high-quality models --
%the focus is usually on model quality, and 
%they search for the best algorithms, features, and hyperparameters~\cite{lopez2023alphad3m}. Pipeline generation for data harmonization can build on such advancements. 
%Still, the task is further complicated by the input data's inherent uncertainty and variability.
%
One challenge for harmonization agents is that to guide the search, we need to design optimizers that measure harmonization success, a non-trivial tasks, and balance multiple objectives including computational costs.
%Also, our system integrates human-in-the-loop interactions to iteratively refine pipelines based on user feedback and domain expertise. 
%An open challenge is determining how to effectively balance automation and human input~\cite{human-in-the-loopdataintegration2017}.

\myparagraph{Acknowledgments}
We thank Brandon Rose and Jataware for supporting the implementation of Harmonia's initial prototype. This work was supported by NSF awards IIS-2106888 and OAC-2411221, and the DARPA
% Automating Scientific Knowledge Extraction and Modeling (ASKEM) program,
ASKEM program
Agreement No. HR0011262087 and the ARPA-H BDF program. The views, opinions, and findings expressed are those of the authors and should not be interpreted as representing the official views or policies of the DARPA, ARPA-H, the U.S. Government, or NSF.

\bibliographystyle{ACM-Reference-Format}
\bibliography{main}


\begin{thebibliography}{77}

%%% the Web address.

\ifx \showCODEN    \undefined \def \showCODEN     #1{\unskip}     \fi
\ifx \showDOI      \undefined \def \showDOI       #1{#1}\fi
\ifx \showISBNx    \undefined \def \showISBNx     #1{\unskip}     \fi
\ifx \showISBNxiii \undefined \def \showISBNxiii  #1{\unskip}     \fi
\ifx \showISSN     \undefined \def \showISSN      #1{\unskip}     \fi
\ifx \showLCCN     \undefined \def \showLCCN      #1{\unskip}     \fi
\ifx \shownote     \undefined \def \shownote      #1{#1}          \fi
\ifx \showarticletitle \undefined \def \showarticletitle #1{#1}   \fi
\ifx \showURL      \undefined \def \showURL       {\relax}        \fi
% invisible to TeX
\providecommand\bibfield[2]{#2}
\providecommand\bibinfo[2]{#2}
\providecommand\natexlab[1]{#1}
\providecommand\showeprint[2][]{arXiv:#2}

\bibitem[\protect\citeauthoryear{Aras, Fuhr, Hwang, de~Keijzer, Klan, Lenz,
  Matth\'{e}, Schweppe, Stern, and De~Tr\'{e}}{Aras et~al\mbox{.}}{2009}]%
        {dagstuhl2029explanation}
\bibfield{author}{\bibinfo{person}{Hidir Aras}, \bibinfo{person}{Norbert Fuhr},
  \bibinfo{person}{Seung-won Hwang}, \bibinfo{person}{Ander de Keijzer},
  \bibinfo{person}{Friederike Klan}, \bibinfo{person}{Hans-Joachim Lenz},
  \bibinfo{person}{Tom Matth\'{e}}, \bibinfo{person}{Heinz Schweppe},
  \bibinfo{person}{Mirco Stern}, {and} \bibinfo{person}{Guy De~Tr\'{e}}.}
  \bibinfo{year}{2009}\natexlab{}.
\newblock \showarticletitle{{08421 Working Group: Explanation}}. In
  \bibinfo{booktitle}{\emph{Uncertainty Management in Information Systems}}
  \emph{(\bibinfo{series}{Dagstuhl Seminar Proceedings (DagSemProc)})},
  \bibfield{editor}{\bibinfo{person}{Christoph Koch}, \bibinfo{person}{Birgitta
  K\"{o}nig-Ries}, \bibinfo{person}{Volker Markl}, {and}
  \bibinfo{person}{Maurice van Keulen}} (Eds.), Vol.~\bibinfo{volume}{8421}.
  \bibinfo{publisher}{Schloss Dagstuhl -- Leibniz-Zentrum f{\"u}r Informatik},
  \bibinfo{address}{Dagstuhl, Germany}, \bibinfo{pages}{1--3}.
\newblock
\showISSN{1862-4405}
\urldef\tempurl%
\url{https://doi.org/10.4230/DagSemProc.08421.4}
\showDOI{\tempurl}


\bibitem[\protect\citeauthoryear{archytas}{archytas}{[n.d.]}]%
        {archytas}
archytas \bibinfo{year}{[n.d.]}\natexlab{}.
\newblock \bibinfo{title}{Archytas: A Tools Interface for AI Agents}.
\newblock \bibinfo{howpublished}{\url{}}.
\newblock


\bibitem[\protect\citeauthoryear{Barrie, Palaiologou, and T{\"o}rnberg}{Barrie
  et~al\mbox{.}}{2024}]%
        {barrie2024prompt}
\bibfield{author}{\bibinfo{person}{Christopher Barrie}, \bibinfo{person}{Elli
  Palaiologou}, {and} \bibinfo{person}{Petter T{\"o}rnberg}.}
  \bibinfo{year}{2024}\natexlab{}.
\newblock \bibinfo{title}{Prompt stability scoring for text annotation with
  large language models}.
\newblock \bibinfo{howpublished}{arXiv preprint arXiv:2407.02039}.
\newblock


\bibitem[\protect\citeauthoryear{bdi-kit}{bdi-kit}{[n.d.]}]%
        {bdi-kit-github}
bdi-kit \bibinfo{year}{[n.d.]}\natexlab{}.
\newblock \bibinfo{title}{The {bdi-kit} data harmonization library}.
\newblock \bibinfo{howpublished}{\url{https://github.com/VIDA-NYU/bdi-kit}}.
\newblock


\bibitem[\protect\citeauthoryear{beaker}{beaker}{[n.d.]}]%
        {beaker}
beaker \bibinfo{year}{[n.d.]}\natexlab{}.
\newblock \bibinfo{title}{Beaker-Kernel: Contextually-aware notebooks with
  built-in AI assistant}.
\newblock
  \bibinfo{howpublished}{\url{https://github.com/jataware/beaker-kernel}}.
\newblock


\bibitem[\protect\citeauthoryear{Berti-Equille}{Berti-Equille}{2019}]%
        {berti2019learn2clean}
\bibfield{author}{\bibinfo{person}{Laure Berti-Equille}.}
  \bibinfo{year}{2019}\natexlab{}.
\newblock \showarticletitle{Learn2clean: Optimizing the sequence of tasks for
  web data preparation}. In \bibinfo{booktitle}{\emph{The world wide web
  conference}}. \bibinfo{pages}{2580--2586}.
\newblock


\bibitem[\protect\citeauthoryear{Cao, Huang, Zhou, Hu, Lih, Savage, Krug,
  Clark, Schnaubelt, Chen, et~al\mbox{.}}{Cao et~al\mbox{.}}{2021}]%
        {cao2021proteogenomic}
\bibfield{author}{\bibinfo{person}{Liwei Cao}, \bibinfo{person}{Chen Huang},
  \bibinfo{person}{Daniel~Cui Zhou}, \bibinfo{person}{Yingwei Hu},
  \bibinfo{person}{T~Mamie Lih}, \bibinfo{person}{Sara~R Savage},
  \bibinfo{person}{Karsten Krug}, \bibinfo{person}{David~J Clark},
  \bibinfo{person}{Michael Schnaubelt}, \bibinfo{person}{Lijun Chen},
  {et~al\mbox{.}}} \bibinfo{year}{2021}\natexlab{}.
\newblock \showarticletitle{Proteogenomic characterization of pancreatic ductal
  adenocarcinoma}.
\newblock \bibinfo{journal}{\emph{Cell}} \bibinfo{volume}{184},
  \bibinfo{number}{19} (\bibinfo{year}{2021}), \bibinfo{pages}{5031--5052}.
\newblock


\bibitem[\protect\citeauthoryear{Chan, Chowdhury, Jaffe, Aung, Sherburn, Mays,
  Starace, Liu, Maksin, Patwardhan, et~al\mbox{.}}{Chan et~al\mbox{.}}{2024}]%
        {chan2024mle}
\bibfield{author}{\bibinfo{person}{Jun~Shern Chan}, \bibinfo{person}{Neil
  Chowdhury}, \bibinfo{person}{Oliver Jaffe}, \bibinfo{person}{James Aung},
  \bibinfo{person}{Dane Sherburn}, \bibinfo{person}{Evan Mays},
  \bibinfo{person}{Giulio Starace}, \bibinfo{person}{Kevin Liu},
  \bibinfo{person}{Leon Maksin}, \bibinfo{person}{Tejal Patwardhan},
  {et~al\mbox{.}}} \bibinfo{year}{2024}\natexlab{}.
\newblock \bibinfo{title}{Mle-bench: Evaluating machine learning agents on
  machine learning engineering}.
\newblock \bibinfo{howpublished}{arXiv preprint arXiv:2410.07095}.
\newblock


\bibitem[\protect\citeauthoryear{Chapman, Missier, Simonelli, and
  Torlone}{Chapman et~al\mbox{.}}{2020}]%
        {chapmanCapturing2020}
\bibfield{author}{\bibinfo{person}{Adriane Chapman}, \bibinfo{person}{Paolo
  Missier}, \bibinfo{person}{Giulia Simonelli}, {and} \bibinfo{person}{Riccardo
  Torlone}.} \bibinfo{year}{2020}\natexlab{}.
\newblock \showarticletitle{Capturing and querying fine-grained provenance of
  preprocessing pipelines in data science}.
\newblock \bibinfo{journal}{\emph{Proc. VLDB Endow.}} \bibinfo{volume}{14},
  \bibinfo{number}{4} (\bibinfo{date}{Dec.} \bibinfo{year}{2020}),
  \bibinfo{pages}{507–520}.
\newblock
\showISSN{2150-8097}
\urldef\tempurl%
\url{https://doi.org/10.14778/3436905.3436911}
\showDOI{\tempurl}


\bibitem[\protect\citeauthoryear{Chen, He, Cui, Fan, Ge, Zhang, Zhang, and
  Chaudhuri}{Chen et~al\mbox{.}}{2024a}]%
        {autoformula2024}
\bibfield{author}{\bibinfo{person}{Sibei Chen}, \bibinfo{person}{Yeye He},
  \bibinfo{person}{Weiwei Cui}, \bibinfo{person}{Ju Fan}, \bibinfo{person}{Song
  Ge}, \bibinfo{person}{Haidong Zhang}, \bibinfo{person}{Dongmei Zhang}, {and}
  \bibinfo{person}{Surajit Chaudhuri}.} \bibinfo{year}{2024}\natexlab{a}.
\newblock \showarticletitle{Auto-Formula: Recommend Formulas in Spreadsheets
  using Contrastive Learning for Table Representations}.
\newblock \bibinfo{journal}{\emph{Proc. ACM Manag. Data}} \bibinfo{volume}{2},
  \bibinfo{number}{3}, Article \bibinfo{articleno}{122} (\bibinfo{date}{May}
  \bibinfo{year}{2024}), \bibinfo{numpages}{27}~pages.
\newblock
\urldef\tempurl%
\url{https://doi.org/10.1145/3654925}
\showDOI{\tempurl}


\bibitem[\protect\citeauthoryear{Chen, Yuan, Zhang, Zheng, Liu, Ni, and
  Hao}{Chen et~al\mbox{.}}{2024b}]%
        {SheetAgent2024}
\bibfield{author}{\bibinfo{person}{Yibin Chen}, \bibinfo{person}{Yifu Yuan},
  \bibinfo{person}{Zeyu Zhang}, \bibinfo{person}{Yan Zheng},
  \bibinfo{person}{Jinyi Liu}, \bibinfo{person}{Fei Ni}, {and}
  \bibinfo{person}{Jianye Hao}.} \bibinfo{year}{2024}\natexlab{b}.
\newblock \showarticletitle{SheetAgent: {A} Generalist Agent for Spreadsheet
  Reasoning and Manipulation via Large Language Models}.
\newblock \bibinfo{journal}{\emph{CoRR}}  \bibinfo{volume}{abs/2403.03636}
  (\bibinfo{year}{2024}).
\newblock
\urldef\tempurl%
\url{https://doi.org/10.48550/ARXIV.2403.03636}
\showDOI{\tempurl}


\bibitem[\protect\citeauthoryear{Cheng, Messerschmidt, Bravo, Waldbauer,
  Bhavikatti, Schenk, Grujic, Model, Kubinec, and Barcel{\'o}}{Cheng
  et~al\mbox{.}}{2024}]%
        {cheng2024general}
\bibfield{author}{\bibinfo{person}{Cindy Cheng}, \bibinfo{person}{Luca
  Messerschmidt}, \bibinfo{person}{Isaac Bravo}, \bibinfo{person}{Marco
  Waldbauer}, \bibinfo{person}{Rohan Bhavikatti}, \bibinfo{person}{Caress
  Schenk}, \bibinfo{person}{Vanja Grujic}, \bibinfo{person}{Tim Model},
  \bibinfo{person}{Robert Kubinec}, {and} \bibinfo{person}{Joan Barcel{\'o}}.}
  \bibinfo{year}{2024}\natexlab{}.
\newblock \showarticletitle{A general primer for data harmonization}.
\newblock \bibinfo{journal}{\emph{Scientific data}} \bibinfo{volume}{11},
  \bibinfo{number}{1} (\bibinfo{year}{2024}), \bibinfo{pages}{152}.
\newblock


\bibitem[\protect\citeauthoryear{Christophides, Efthymiou, Palpanas, Papadakis,
  and Stefanidis}{Christophides et~al\mbox{.}}{2020}]%
        {christophides2020overview}
\bibfield{author}{\bibinfo{person}{Vassilis Christophides},
  \bibinfo{person}{Vasilis Efthymiou}, \bibinfo{person}{Themis Palpanas},
  \bibinfo{person}{George Papadakis}, {and} \bibinfo{person}{Kostas
  Stefanidis}.} \bibinfo{year}{2020}\natexlab{}.
\newblock \showarticletitle{An overview of end-to-end entity resolution for big
  data}.
\newblock \bibinfo{journal}{\emph{ACM Computing Surveys (CSUR)}}
  \bibinfo{volume}{53}, \bibinfo{number}{6} (\bibinfo{year}{2020}),
  \bibinfo{pages}{1--42}.
\newblock


\bibitem[\protect\citeauthoryear{Clark, Dhanasekaran, Petralia, Pan, Song, Hu,
  da~Veiga~Leprevost, Reva, Lih, Chang, et~al\mbox{.}}{Clark
  et~al\mbox{.}}{2019}]%
        {clark2019integrated}
\bibfield{author}{\bibinfo{person}{David~J Clark}, \bibinfo{person}{Saravana~M
  Dhanasekaran}, \bibinfo{person}{Francesca Petralia}, \bibinfo{person}{Jianbo
  Pan}, \bibinfo{person}{Xiaoyu Song}, \bibinfo{person}{Yingwei Hu},
  \bibinfo{person}{Felipe da Veiga~Leprevost}, \bibinfo{person}{Boris Reva},
  \bibinfo{person}{Tung-Shing~M Lih}, \bibinfo{person}{Hui-Yin Chang},
  {et~al\mbox{.}}} \bibinfo{year}{2019}\natexlab{}.
\newblock \showarticletitle{Integrated proteogenomic characterization of clear
  cell renal cell carcinoma}.
\newblock \bibinfo{journal}{\emph{Cell}} \bibinfo{volume}{179},
  \bibinfo{number}{4} (\bibinfo{year}{2019}), \bibinfo{pages}{964--983}.
\newblock


\bibitem[\protect\citeauthoryear{Dargahi~Nobari and Rafiei}{Dargahi~Nobari and
  Rafiei}{2024}]%
        {dtt2024}
\bibfield{author}{\bibinfo{person}{Arash Dargahi~Nobari} {and}
  \bibinfo{person}{Davood Rafiei}.} \bibinfo{year}{2024}\natexlab{}.
\newblock \showarticletitle{DTT: An Example-Driven Tabular Transformer for
  Joinability by Leveraging Large Language Models}.
\newblock \bibinfo{journal}{\emph{Proc. ACM Manag. Data}} \bibinfo{volume}{2},
  \bibinfo{number}{1}, Article \bibinfo{articleno}{24} (\bibinfo{date}{March}
  \bibinfo{year}{2024}), \bibinfo{numpages}{24}~pages.
\newblock
\urldef\tempurl%
\url{https://doi.org/10.1145/3639279}
\showDOI{\tempurl}


\bibitem[\protect\citeauthoryear{Dong, Halevy, and Yu}{Dong
  et~al\mbox{.}}{2009}]%
        {uncertainty2009}
\bibfield{author}{\bibinfo{person}{Xin~Luna Dong}, \bibinfo{person}{Alon
  Halevy}, {and} \bibinfo{person}{Cong Yu}.} \bibinfo{year}{2009}\natexlab{}.
\newblock \showarticletitle{Data integration with uncertainty}.
\newblock \bibinfo{journal}{\emph{The VLDB Journal}} \bibinfo{volume}{18},
  \bibinfo{number}{2} (\bibinfo{date}{April} \bibinfo{year}{2009}),
  \bibinfo{pages}{469–500}.
\newblock
\showISSN{1066-8888}
\urldef\tempurl%
\url{https://doi.org/10.1007/s00778-008-0119-9}
\showDOI{\tempurl}


\bibitem[\protect\citeauthoryear{Dou, Katsnelson, Gritsenko, Hu, Reva, Hong,
  Wang, Kolodziejczak, Lu, Tsai, et~al\mbox{.}}{Dou et~al\mbox{.}}{2023}]%
        {dou2023proteogenomic}
\bibfield{author}{\bibinfo{person}{Yongchao Dou}, \bibinfo{person}{Lizabeth
  Katsnelson}, \bibinfo{person}{Marina~A Gritsenko}, \bibinfo{person}{Yingwei
  Hu}, \bibinfo{person}{Boris Reva}, \bibinfo{person}{Runyu Hong},
  \bibinfo{person}{Yi-Ting Wang}, \bibinfo{person}{Iga Kolodziejczak},
  \bibinfo{person}{Rita Jui-Hsien Lu}, \bibinfo{person}{Chia-Feng Tsai},
  {et~al\mbox{.}}} \bibinfo{year}{2023}\natexlab{}.
\newblock \showarticletitle{Proteogenomic insights suggest druggable pathways
  in endometrial carcinoma}.
\newblock \bibinfo{journal}{\emph{Cancer cell}} \bibinfo{volume}{41},
  \bibinfo{number}{9} (\bibinfo{year}{2023}), \bibinfo{pages}{1586--1605}.
\newblock


\bibitem[\protect\citeauthoryear{Dou, Kawaler, Zhou, Gritsenko, Huang,
  Blumenberg, Karpova, Petyuk, Savage, Satpathy, et~al\mbox{.}}{Dou
  et~al\mbox{.}}{2020}]%
        {dou2020proteogenomic}
\bibfield{author}{\bibinfo{person}{Yongchao Dou}, \bibinfo{person}{Emily~A
  Kawaler}, \bibinfo{person}{Daniel~Cui Zhou}, \bibinfo{person}{Marina~A
  Gritsenko}, \bibinfo{person}{Chen Huang}, \bibinfo{person}{Lili Blumenberg},
  \bibinfo{person}{Alla Karpova}, \bibinfo{person}{Vladislav~A Petyuk},
  \bibinfo{person}{Sara~R Savage}, \bibinfo{person}{Shankha Satpathy},
  {et~al\mbox{.}}} \bibinfo{year}{2020}\natexlab{}.
\newblock \showarticletitle{Proteogenomic characterization of endometrial
  carcinoma}.
\newblock \bibinfo{journal}{\emph{Cell}} \bibinfo{volume}{180},
  \bibinfo{number}{4} (\bibinfo{year}{2020}), \bibinfo{pages}{729--748}.
\newblock


\bibitem[\protect\citeauthoryear{Esfandiarpoor and Bach}{Esfandiarpoor and
  Bach}{2024}]%
        {esfandiarpoor2024followup}
\bibfield{author}{\bibinfo{person}{Reza Esfandiarpoor} {and}
  \bibinfo{person}{Stephen Bach}.} \bibinfo{year}{2024}\natexlab{}.
\newblock \showarticletitle{Follow-Up Differential Descriptions: Language
  Models Resolve Ambiguities for Image Classification}. In
  \bibinfo{booktitle}{\emph{International Conference on Learning
  Representations (ICLR)}}.
\newblock
\urldef\tempurl%
\url{https://openreview.net/forum?id=g6rZtxaXRm}
\showURL{%
\tempurl}


\bibitem[\protect\citeauthoryear{Fan, Han, Fan, Chai, Tang, Li, and Du}{Fan
  et~al\mbox{.}}{2024}]%
        {fan2024cost}
\bibfield{author}{\bibinfo{person}{Meihao Fan}, \bibinfo{person}{Xiaoyue Han},
  \bibinfo{person}{Ju Fan}, \bibinfo{person}{Chengliang Chai},
  \bibinfo{person}{Nan Tang}, \bibinfo{person}{Guoliang Li}, {and}
  \bibinfo{person}{Xiaoyong Du}.} \bibinfo{year}{2024}\natexlab{}.
\newblock \showarticletitle{Cost-effective in-context learning for entity
  resolution: A design space exploration}. In \bibinfo{booktitle}{\emph{2024
  IEEE 40th International Conference on Data Engineering (ICDE)}}. IEEE,
  \bibinfo{pages}{3696--3709}.
\newblock


\bibitem[\protect\citeauthoryear{Fernandez, Elmore, Franklin, Krishnan, and
  Tan}{Fernandez et~al\mbox{.}}{2023}]%
        {llmDisruptVLDB2023}
\bibfield{author}{\bibinfo{person}{Raul~Castro Fernandez},
  \bibinfo{person}{Aaron~J. Elmore}, \bibinfo{person}{Michael~J. Franklin},
  \bibinfo{person}{Sanjay Krishnan}, {and} \bibinfo{person}{Chenhao Tan}.}
  \bibinfo{year}{2023}\natexlab{}.
\newblock \showarticletitle{How Large Language Models Will Disrupt Data
  Management}.
\newblock \bibinfo{journal}{\emph{PVLDB}} \bibinfo{volume}{16},
  \bibinfo{number}{11} (\bibinfo{date}{July} \bibinfo{year}{2023}),
  \bibinfo{pages}{3302–3309}.
\newblock
\showISSN{2150-8097}
\urldef\tempurl%
\url{https://doi.org/10.14778/3611479.3611527}
\showDOI{\tempurl}


\bibitem[\protect\citeauthoryear{Feuer, Liu, Hegde, and Freire}{Feuer
  et~al\mbox{.}}{2024}]%
        {feuer:vldb2024}
\bibfield{author}{\bibinfo{person}{Benjamin Feuer}, \bibinfo{person}{Yurong
  Liu}, \bibinfo{person}{Chinmay Hegde}, {and} \bibinfo{person}{Juliana
  Freire}.} \bibinfo{year}{2024}\natexlab{}.
\newblock \showarticletitle{ArcheType: {A} Novel Framework for Open-Source
  Column Type Annotation using Large Language Models}.
\newblock \bibinfo{journal}{\emph{Proc. {VLDB} Endow.}} \bibinfo{volume}{17},
  \bibinfo{number}{9} (\bibinfo{year}{2024}), \bibinfo{pages}{2279--2292}.
\newblock
\urldef\tempurl%
\url{https://www.vldb.org/pvldb/vol17/p2279-freire.pdf}
\showURL{%
\tempurl}


\bibitem[\protect\citeauthoryear{Gillette, Satpathy, Cao, Dhanasekaran,
  Vasaikar, Krug, Petralia, Li, Liang, Reva, et~al\mbox{.}}{Gillette
  et~al\mbox{.}}{2020}]%
        {gillette2020proteogenomic}
\bibfield{author}{\bibinfo{person}{Michael~A Gillette},
  \bibinfo{person}{Shankha Satpathy}, \bibinfo{person}{Song Cao},
  \bibinfo{person}{Saravana~M Dhanasekaran}, \bibinfo{person}{Suhas~V
  Vasaikar}, \bibinfo{person}{Karsten Krug}, \bibinfo{person}{Francesca
  Petralia}, \bibinfo{person}{Yize Li}, \bibinfo{person}{Wen-Wei Liang},
  \bibinfo{person}{Boris Reva}, {et~al\mbox{.}}}
  \bibinfo{year}{2020}\natexlab{}.
\newblock \showarticletitle{Proteogenomic characterization reveals therapeutic
  vulnerabilities in lung adenocarcinoma}.
\newblock \bibinfo{journal}{\emph{Cell}} \bibinfo{volume}{182},
  \bibinfo{number}{1} (\bibinfo{year}{2020}), \bibinfo{pages}{200--225}.
\newblock


\bibitem[\protect\citeauthoryear{Harmonia}{Harmonia}{[n.d.]}]%
        {harmonia-github}
Harmonia \bibinfo{year}{[n.d.]}\natexlab{}.
\newblock \bibinfo{title}{Harmonia: An Interactive Data Harmonization Agent}.
\newblock \bibinfo{howpublished}{\url{https://github.com/VIDA-NYU/harmonia/}}.
\newblock


\bibitem[\protect\citeauthoryear{Harmonia Demonstration.}{Harmonia
  Demonstration.}{[n.d.]}]%
        {harmonia-video}
Harmonia Demonstration. \bibinfo{year}{[n.d.]}\natexlab{}.
\newblock \bibinfo{title}{Harmonia: Interactive Data Harmonization with LLM
  Agents (YouTube)}.
\newblock
  \bibinfo{howpublished}{\url{https://www.youtube.com/watch?v=D25x0B_xs3c}}.
\newblock


\bibitem[\protect\citeauthoryear{Hsieh, Li, Yeh, Nakhost, Fujii, Ratner,
  Krishna, Lee, and Pfister}{Hsieh et~al\mbox{.}}{2023}]%
        {hsieh-etal-2023-distilling}
\bibfield{author}{\bibinfo{person}{Cheng-Yu Hsieh}, \bibinfo{person}{Chun-Liang
  Li}, \bibinfo{person}{Chih-kuan Yeh}, \bibinfo{person}{Hootan Nakhost},
  \bibinfo{person}{Yasuhisa Fujii}, \bibinfo{person}{Alex Ratner},
  \bibinfo{person}{Ranjay Krishna}, \bibinfo{person}{Chen-Yu Lee}, {and}
  \bibinfo{person}{Tomas Pfister}.} \bibinfo{year}{2023}\natexlab{}.
\newblock \showarticletitle{Distilling Step-by-Step! Outperforming Larger
  Language Models with Less Training Data and Smaller Model Sizes}. In
  \bibinfo{booktitle}{\emph{Findings of the Association for Computational
  Linguistics: ACL 2023}}, \bibfield{editor}{\bibinfo{person}{Anna Rogers},
  \bibinfo{person}{Jordan Boyd-Graber}, {and} \bibinfo{person}{Naoaki Okazaki}}
  (Eds.). \bibinfo{publisher}{Association for Computational Linguistics},
  \bibinfo{address}{Toronto, Canada}, \bibinfo{pages}{8003--8017}.
\newblock
\urldef\tempurl%
\url{https://doi.org/10.18653/v1/2023.findings-acl.507}
\showDOI{\tempurl}


\bibitem[\protect\citeauthoryear{Hu, Zhao, Wei, Chai, Ma, Wang, Wang, Su, Xu,
  Zhu, et~al\mbox{.}}{Hu et~al\mbox{.}}{2024}]%
        {hu2024infiagent}
\bibfield{author}{\bibinfo{person}{Xueyu Hu}, \bibinfo{person}{Ziyu Zhao},
  \bibinfo{person}{Shuang Wei}, \bibinfo{person}{Ziwei Chai},
  \bibinfo{person}{Qianli Ma}, \bibinfo{person}{Guoyin Wang},
  \bibinfo{person}{Xuwu Wang}, \bibinfo{person}{Jing Su},
  \bibinfo{person}{Jingjing Xu}, \bibinfo{person}{Ming Zhu}, {et~al\mbox{.}}}
  \bibinfo{year}{2024}\natexlab{}.
\newblock \bibinfo{title}{Infiagent-dabench: Evaluating agents on data analysis
  tasks}.
\newblock \bibinfo{howpublished}{arXiv preprint arXiv:2401.05507}.
\newblock


\bibitem[\protect\citeauthoryear{Huang, Chen, Savage, Eguez, Dou, Li,
  da~Veiga~Leprevost, Jaehnig, Lei, Wen, et~al\mbox{.}}{Huang
  et~al\mbox{.}}{2021}]%
        {huang2021proteogenomic}
\bibfield{author}{\bibinfo{person}{Chen Huang}, \bibinfo{person}{Lijun Chen},
  \bibinfo{person}{Sara~R Savage}, \bibinfo{person}{Rodrigo~Vargas Eguez},
  \bibinfo{person}{Yongchao Dou}, \bibinfo{person}{Yize Li},
  \bibinfo{person}{Felipe da Veiga~Leprevost}, \bibinfo{person}{Eric~J
  Jaehnig}, \bibinfo{person}{Jonathan~T Lei}, \bibinfo{person}{Bo Wen},
  {et~al\mbox{.}}} \bibinfo{year}{2021}\natexlab{}.
\newblock \showarticletitle{Proteogenomic insights into the biology and
  treatment of HPV-negative head and neck squamous cell carcinoma}.
\newblock \bibinfo{journal}{\emph{Cancer cell}} \bibinfo{volume}{39},
  \bibinfo{number}{3} (\bibinfo{year}{2021}), \bibinfo{pages}{361--379}.
\newblock


\bibitem[\protect\citeauthoryear{Huang, Vora, Liang, and Leskovec}{Huang
  et~al\mbox{.}}{2024b}]%
        {huang2024mlagentbench}
\bibfield{author}{\bibinfo{person}{Qian Huang}, \bibinfo{person}{Jian Vora},
  \bibinfo{person}{Percy Liang}, {and} \bibinfo{person}{Jure Leskovec}.}
  \bibinfo{year}{2024}\natexlab{b}.
\newblock \showarticletitle{{MLAgentBench: Evaluating Language Agents on
  Machine Learning Experimentation}}. In
  \bibinfo{booktitle}{\emph{International Conference on Machine Learning}}.
\newblock


\bibitem[\protect\citeauthoryear{Huang, Sun, Wang, Wu, Zhang, Li, Gao, Huang,
  Lyu, Zhang, Li, Sun, Liu, Liu, Wang, Zhang, Vidgen, Kailkhura, Xiong, Xiao,
  Li, Xing, Huang, Liu, Ji, Wang, Zhang, Yao, Kellis, Zitnik, Jiang, Bansal,
  Zou, Pei, Liu, Gao, Han, Zhao, Tang, Wang, Vanschoren, Mitchell, Shu, Xu,
  Chang, He, Huang, Backes, Gong, Yu, Chen, Gu, Xu, Ying, Ji, Jana, Chen, Liu,
  Zhou, Wang, Li, Zhang, Wang, Xie, Chen, Wang, Liu, Ye, Cao, Chen, and
  Zhao}{Huang et~al\mbox{.}}{2024a}]%
        {pmlr-v235-huang24x}
\bibfield{author}{\bibinfo{person}{Yue Huang}, \bibinfo{person}{Lichao Sun},
  \bibinfo{person}{Haoran Wang}, \bibinfo{person}{Siyuan Wu},
  \bibinfo{person}{Qihui Zhang}, \bibinfo{person}{Yuan Li},
  \bibinfo{person}{Chujie Gao}, \bibinfo{person}{Yixin Huang},
  \bibinfo{person}{Wenhan Lyu}, \bibinfo{person}{Yixuan Zhang},
  \bibinfo{person}{Xiner Li}, \bibinfo{person}{Hanchi Sun},
  \bibinfo{person}{Zhengliang Liu}, \bibinfo{person}{Yixin Liu},
  \bibinfo{person}{Yijue Wang}, \bibinfo{person}{Zhikun Zhang},
  \bibinfo{person}{Bertie Vidgen}, \bibinfo{person}{Bhavya Kailkhura},
  \bibinfo{person}{Caiming Xiong}, \bibinfo{person}{Chaowei Xiao},
  \bibinfo{person}{Chunyuan Li}, \bibinfo{person}{Eric~P. Xing},
  \bibinfo{person}{Furong Huang}, \bibinfo{person}{Hao Liu},
  \bibinfo{person}{Heng Ji}, \bibinfo{person}{Hongyi Wang},
  \bibinfo{person}{Huan Zhang}, \bibinfo{person}{Huaxiu Yao},
  \bibinfo{person}{Manolis Kellis}, \bibinfo{person}{Marinka Zitnik},
  \bibinfo{person}{Meng Jiang}, \bibinfo{person}{Mohit Bansal},
  \bibinfo{person}{James Zou}, \bibinfo{person}{Jian Pei},
  \bibinfo{person}{Jian Liu}, \bibinfo{person}{Jianfeng Gao},
  \bibinfo{person}{Jiawei Han}, \bibinfo{person}{Jieyu Zhao},
  \bibinfo{person}{Jiliang Tang}, \bibinfo{person}{Jindong Wang},
  \bibinfo{person}{Joaquin Vanschoren}, \bibinfo{person}{John Mitchell},
  \bibinfo{person}{Kai Shu}, \bibinfo{person}{Kaidi Xu},
  \bibinfo{person}{Kai-Wei Chang}, \bibinfo{person}{Lifang He},
  \bibinfo{person}{Lifu Huang}, \bibinfo{person}{Michael Backes},
  \bibinfo{person}{Neil~Zhenqiang Gong}, \bibinfo{person}{Philip~S. Yu},
  \bibinfo{person}{Pin-Yu Chen}, \bibinfo{person}{Quanquan Gu},
  \bibinfo{person}{Ran Xu}, \bibinfo{person}{Rex Ying},
  \bibinfo{person}{Shuiwang Ji}, \bibinfo{person}{Suman Jana},
  \bibinfo{person}{Tianlong Chen}, \bibinfo{person}{Tianming Liu},
  \bibinfo{person}{Tianyi Zhou}, \bibinfo{person}{William~Yang Wang},
  \bibinfo{person}{Xiang Li}, \bibinfo{person}{Xiangliang Zhang},
  \bibinfo{person}{Xiao Wang}, \bibinfo{person}{Xing Xie}, \bibinfo{person}{Xun
  Chen}, \bibinfo{person}{Xuyu Wang}, \bibinfo{person}{Yan Liu},
  \bibinfo{person}{Yanfang Ye}, \bibinfo{person}{Yinzhi Cao},
  \bibinfo{person}{Yong Chen}, {and} \bibinfo{person}{Yue Zhao}.}
  \bibinfo{year}{2024}\natexlab{a}.
\newblock \showarticletitle{Position: {T}rust{LLM}: Trustworthiness in Large
  Language Models}. In \bibinfo{booktitle}{\emph{Proceedings of the 41st
  International Conference on Machine Learning}}
  \emph{(\bibinfo{series}{Proceedings of Machine Learning Research})},
  \bibfield{editor}{\bibinfo{person}{Ruslan Salakhutdinov},
  \bibinfo{person}{Zico Kolter}, \bibinfo{person}{Katherine Heller},
  \bibinfo{person}{Adrian Weller}, \bibinfo{person}{Nuria Oliver},
  \bibinfo{person}{Jonathan Scarlett}, {and} \bibinfo{person}{Felix
  Berkenkamp}} (Eds.), Vol.~\bibinfo{volume}{235}. \bibinfo{publisher}{PMLR},
  \bibinfo{pages}{20166--20270}.
\newblock
\urldef\tempurl%
\url{https://proceedings.mlr.press/v235/huang24x.html}
\showURL{%
\tempurl}


\bibitem[\protect\citeauthoryear{Institute}{Institute}{2024}]%
        {gdc}
\bibfield{author}{\bibinfo{person}{National~Cancer Institute}.}
  \bibinfo{year}{2024}\natexlab{}.
\newblock \bibinfo{title}{GDC Data Model}.
\newblock
  \bibinfo{howpublished}{\url{https://gdc.cancer.gov/developers/gdc-data-model}}.
\newblock
\newblock
\shownote{Accessed: 2024-09-20.}


\bibitem[\protect\citeauthoryear{Jin, Han, Yang, Jiang, Liu, Chang, Chen, and
  Hu}{Jin et~al\mbox{.}}{2024}]%
        {jin2024llm}
\bibfield{author}{\bibinfo{person}{Hongye Jin}, \bibinfo{person}{Xiaotian Han},
  \bibinfo{person}{Jingfeng Yang}, \bibinfo{person}{Zhimeng Jiang},
  \bibinfo{person}{Zirui Liu}, \bibinfo{person}{Chia-Yuan Chang},
  \bibinfo{person}{Huiyuan Chen}, {and} \bibinfo{person}{Xia Hu}.}
  \bibinfo{year}{2024}\natexlab{}.
\newblock \bibinfo{title}{Llm maybe longlm: Self-extend llm context window
  without tuning}.
\newblock \bibinfo{howpublished}{arXiv preprint arXiv:2401.01325}.
\newblock


\bibitem[\protect\citeauthoryear{jupyter}{jupyter}{[n.d.]}]%
        {jupyter}
jupyter \bibinfo{year}{[n.d.]}\natexlab{}.
\newblock \bibinfo{title}{Project Jupyter}.
\newblock \bibinfo{howpublished}{\url{https://jupyter.org/}}.
\newblock


\bibitem[\protect\citeauthoryear{Kayali, Lykov, Fountalis, Vasiloglou, Olteanu,
  and Suciu}{Kayali et~al\mbox{.}}{2024}]%
        {chorus-vldb2024}
\bibfield{author}{\bibinfo{person}{Moe Kayali}, \bibinfo{person}{Anton Lykov},
  \bibinfo{person}{Ilias Fountalis}, \bibinfo{person}{Nikolaos Vasiloglou},
  \bibinfo{person}{Dan Olteanu}, {and} \bibinfo{person}{Dan Suciu}.}
  \bibinfo{year}{2024}\natexlab{}.
\newblock \showarticletitle{{CHORUS:} Foundation Models for Unified Data
  Discovery and Exploration}.
\newblock \bibinfo{journal}{\emph{Proc. {VLDB} Endow.}} \bibinfo{volume}{17},
  \bibinfo{number}{8} (\bibinfo{year}{2024}), \bibinfo{pages}{2104--2114}.
\newblock
\urldef\tempurl%
\url{https://www.vldb.org/pvldb/vol17/p2104-kayali.pdf}
\showURL{%
\tempurl}


\bibitem[\protect\citeauthoryear{Khattab, Singhvi, Maheshwari, Zhang,
  Santhanam, A, Haq, Sharma, Joshi, Moazam, Miller, Zaharia, and Potts}{Khattab
  et~al\mbox{.}}{2024}]%
        {khattab2024dspy}
\bibfield{author}{\bibinfo{person}{Omar Khattab}, \bibinfo{person}{Arnav
  Singhvi}, \bibinfo{person}{Paridhi Maheshwari}, \bibinfo{person}{Zhiyuan
  Zhang}, \bibinfo{person}{Keshav Santhanam}, \bibinfo{person}{Sri~Vardhamanan
  A}, \bibinfo{person}{Saiful Haq}, \bibinfo{person}{Ashutosh Sharma},
  \bibinfo{person}{Thomas~T. Joshi}, \bibinfo{person}{Hanna Moazam},
  \bibinfo{person}{Heather Miller}, \bibinfo{person}{Matei Zaharia}, {and}
  \bibinfo{person}{Christopher Potts}.} \bibinfo{year}{2024}\natexlab{}.
\newblock \showarticletitle{{DSP}y: Compiling Declarative Language Model Calls
  into State-of-the-Art Pipelines}. In \bibinfo{booktitle}{\emph{The Twelfth
  International Conference on Learning Representations}}.
\newblock
\urldef\tempurl%
\url{https://openreview.net/forum?id=sY5N0zY5Od}
\showURL{%
\tempurl}


\bibitem[\protect\citeauthoryear{Kolaitis}{Kolaitis}{2018}]%
        {dataexchange2018}
\bibfield{author}{\bibinfo{person}{Phokion~G. Kolaitis}.}
  \bibinfo{year}{2018}\natexlab{}.
\newblock \showarticletitle{Reflections on Schema Mappings, Data Exchange, and
  Metadata Management}. In \bibinfo{booktitle}{\emph{Proceedings of the 37th
  ACM SIGMOD-SIGACT-SIGAI Symposium on Principles of Database Systems}}
  (Houston, TX, USA) \emph{(\bibinfo{series}{PODS '18})}.
  \bibinfo{publisher}{Association for Computing Machinery},
  \bibinfo{address}{New York, NY, USA}, \bibinfo{pages}{107–109}.
\newblock
\showISBNx{9781450347068}
\urldef\tempurl%
\url{https://doi.org/10.1145/3196959.3196991}
\showDOI{\tempurl}


\bibitem[\protect\citeauthoryear{Koop, Scheidegger, Callahan, Freire, and
  Silva}{Koop et~al\mbox{.}}{2008}]%
        {koop2008viscomplete}
\bibfield{author}{\bibinfo{person}{David Koop}, \bibinfo{person}{Carlos~E
  Scheidegger}, \bibinfo{person}{Steven~P Callahan}, \bibinfo{person}{Juliana
  Freire}, {and} \bibinfo{person}{Cl{\'a}udio~T Silva}.}
  \bibinfo{year}{2008}\natexlab{}.
\newblock \showarticletitle{Viscomplete: Automating suggestions for
  visualization pipelines}.
\newblock \bibinfo{journal}{\emph{IEEE Transactions on Visualization and
  Computer Graphics}} \bibinfo{volume}{14}, \bibinfo{number}{6}
  (\bibinfo{year}{2008}), \bibinfo{pages}{1691--1698}.
\newblock


\bibitem[\protect\citeauthoryear{Koutras, Siachamis, Ionescu, Psarakis, Brons,
  Fragkoulis, Lofi, Bonifati, and Katsifodimos}{Koutras et~al\mbox{.}}{2021}]%
        {koutras2021valentine}
\bibfield{author}{\bibinfo{person}{Christos Koutras}, \bibinfo{person}{George
  Siachamis}, \bibinfo{person}{Andra Ionescu}, \bibinfo{person}{Kyriakos
  Psarakis}, \bibinfo{person}{Jerry Brons}, \bibinfo{person}{Marios
  Fragkoulis}, \bibinfo{person}{Christoph Lofi}, \bibinfo{person}{Angela
  Bonifati}, {and} \bibinfo{person}{Asterios Katsifodimos}.}
  \bibinfo{year}{2021}\natexlab{}.
\newblock \showarticletitle{Valentine: Evaluating matching techniques for
  dataset discovery}. In \bibinfo{booktitle}{\emph{2021 IEEE 37th International
  Conference on Data Engineering (ICDE)}}. IEEE, \bibinfo{pages}{468--479}.
\newblock


\bibitem[\protect\citeauthoryear{Krug, Jaehnig, Satpathy, Blumenberg, Karpova,
  Anurag, Miles, Mertins, Geffen, Tang, et~al\mbox{.}}{Krug
  et~al\mbox{.}}{2020}]%
        {krug2020proteogenomic}
\bibfield{author}{\bibinfo{person}{Karsten Krug}, \bibinfo{person}{Eric~J
  Jaehnig}, \bibinfo{person}{Shankha Satpathy}, \bibinfo{person}{Lili
  Blumenberg}, \bibinfo{person}{Alla Karpova}, \bibinfo{person}{Meenakshi
  Anurag}, \bibinfo{person}{George Miles}, \bibinfo{person}{Philipp Mertins},
  \bibinfo{person}{Yifat Geffen}, \bibinfo{person}{Lauren~C Tang},
  {et~al\mbox{.}}} \bibinfo{year}{2020}\natexlab{}.
\newblock \showarticletitle{Proteogenomic landscape of breast cancer
  tumorigenesis and targeted therapy}.
\newblock \bibinfo{journal}{\emph{Cell}} \bibinfo{volume}{183},
  \bibinfo{number}{5} (\bibinfo{year}{2020}), \bibinfo{pages}{1436--1456}.
\newblock


\bibitem[\protect\citeauthoryear{langchain}{langchain}{[n.d.]}]%
        {langchain}
langchain \bibinfo{year}{[n.d.]}\natexlab{}.
\newblock \bibinfo{title}{LangChain}.
\newblock \bibinfo{howpublished}{\url{https://www.langchain.com/}}.
\newblock


\bibitem[\protect\citeauthoryear{Li, Hammoud, Itani, Khizbullin, and Ghanem}{Li
  et~al\mbox{.}}{2023b}]%
        {li2023camel}
\bibfield{author}{\bibinfo{person}{Guohao Li}, \bibinfo{person}{Hasan Hammoud},
  \bibinfo{person}{Hani Itani}, \bibinfo{person}{Dmitrii Khizbullin}, {and}
  \bibinfo{person}{Bernard Ghanem}.} \bibinfo{year}{2023}\natexlab{b}.
\newblock \showarticletitle{Camel: Communicative agents for" mind" exploration
  of large language model society}.
\newblock \bibinfo{journal}{\emph{Advances in Neural Information Processing
  Systems}}  \bibinfo{volume}{36} (\bibinfo{year}{2023}),
  \bibinfo{pages}{51991--52008}.
\newblock


\bibitem[\protect\citeauthoryear{Li, He, Yan, Wang, and Chaudhuri}{Li
  et~al\mbox{.}}{2023c}]%
        {autotables2023}
\bibfield{author}{\bibinfo{person}{Peng Li}, \bibinfo{person}{Yeye He},
  \bibinfo{person}{Cong Yan}, \bibinfo{person}{Yue Wang}, {and}
  \bibinfo{person}{Surajit Chaudhuri}.} \bibinfo{year}{2023}\natexlab{c}.
\newblock \showarticletitle{Auto-Tables: Synthesizing Multi-Step
  Transformations to Relationalize Tables without Using Examples}.
\newblock \bibinfo{journal}{\emph{PVLDB}} \bibinfo{volume}{16},
  \bibinfo{number}{11} (\bibinfo{date}{July} \bibinfo{year}{2023}),
  \bibinfo{pages}{3391–3403}.
\newblock
\showISSN{2150-8097}
\urldef\tempurl%
\url{https://doi.org/10.14778/3611479.3611534}
\showDOI{\tempurl}


\bibitem[\protect\citeauthoryear{Li, Dou, Leprevost, Geffen, Calinawan, Aguet,
  Akiyama, Anand, Birger, Cao, et~al\mbox{.}}{Li et~al\mbox{.}}{2023a}]%
        {li2023proteogenomic}
\bibfield{author}{\bibinfo{person}{Yize Li}, \bibinfo{person}{Yongchao Dou},
  \bibinfo{person}{Felipe Da~Veiga Leprevost}, \bibinfo{person}{Yifat Geffen},
  \bibinfo{person}{Anna~P Calinawan}, \bibinfo{person}{Fran{\c{c}}ois Aguet},
  \bibinfo{person}{Yo Akiyama}, \bibinfo{person}{Shankara Anand},
  \bibinfo{person}{Chet Birger}, \bibinfo{person}{Song Cao}, {et~al\mbox{.}}}
  \bibinfo{year}{2023}\natexlab{a}.
\newblock \showarticletitle{Proteogenomic data and resources for pan-cancer
  analysis}.
\newblock \bibinfo{journal}{\emph{Cancer cell}} \bibinfo{volume}{41},
  \bibinfo{number}{8} (\bibinfo{year}{2023}), \bibinfo{pages}{1397--1406}.
\newblock


\bibitem[\protect\citeauthoryear{Liu, Pena, Santos, Wu, and Freire}{Liu
  et~al\mbox{.}}{2025}]%
        {magneto-vldb2025}
\bibfield{author}{\bibinfo{person}{Yurong Liu}, \bibinfo{person}{Eduardo Pena},
  \bibinfo{person}{Aecio Santos}, \bibinfo{person}{Eden Wu}, {and}
  \bibinfo{person}{Juliana Freire}.} \bibinfo{year}{2025}\natexlab{}.
\newblock \showarticletitle{Magneto: Combining Small and Large Language Models
  for Schema Matching}.
\newblock \bibinfo{journal}{\emph{Proceedings of the VLDB Endowment}}
  \bibinfo{volume}{18}, \bibinfo{number}{8} (\bibinfo{year}{2025}),
  \bibinfo{pages}{2681--2694}.
\newblock
\urldef\tempurl%
\url{https://doi.org/10.14778/3742728.3742757}
\showDOI{\tempurl}


\bibitem[\protect\citeauthoryear{Liu, Santos, Pena, Lopez, Wu, and Freire}{Liu
  et~al\mbox{.}}{2024}]%
        {liu2024enhancing}
\bibfield{author}{\bibinfo{person}{Yurong Liu}, \bibinfo{person}{A{\'e}cio
  Santos}, \bibinfo{person}{Eduardo~HM Pena}, \bibinfo{person}{Roque Lopez},
  \bibinfo{person}{Eden Wu}, {and} \bibinfo{person}{Juliana Freire}.}
  \bibinfo{year}{2024}\natexlab{}.
\newblock \showarticletitle{Enhancing Biomedical Schema Matching with LLM-based
  Training Data Generation}. In \bibinfo{booktitle}{\emph{NeurIPS 2024 Third
  Table Representation Learning Workshop}}.
\newblock


\bibitem[\protect\citeauthoryear{Lopez, Louren{\c{c}}o, Rampin, Castelo,
  Santos, Ono, Silva, and Freire}{Lopez et~al\mbox{.}}{2023}]%
        {lopez2023alphad3m}
\bibfield{author}{\bibinfo{person}{Roque Lopez}, \bibinfo{person}{Raoni
  Louren{\c{c}}o}, \bibinfo{person}{Remi Rampin}, \bibinfo{person}{Sonia
  Castelo}, \bibinfo{person}{A{\'e}cio~SR Santos}, \bibinfo{person}{Jorge
  Henrique~Piazentin Ono}, \bibinfo{person}{Claudio Silva}, {and}
  \bibinfo{person}{Juliana Freire}.} \bibinfo{year}{2023}\natexlab{}.
\newblock \showarticletitle{AlphaD3M: An Open-Source AutoML Library for
  Multiple ML Tasks}. In \bibinfo{booktitle}{\emph{International Conference on
  Automated Machine Learning}}. PMLR, \bibinfo{pages}{22--1}.
\newblock


\bibitem[\protect\citeauthoryear{Ma, Yang, Xiong, Chen, Yu, Zhang, May,
  Zettlemoyer, Levy, and Zhou}{Ma et~al\mbox{.}}{2024a}]%
        {ma2024megalodon}
\bibfield{author}{\bibinfo{person}{Xuezhe Ma}, \bibinfo{person}{Xiaomeng Yang},
  \bibinfo{person}{Wenhan Xiong}, \bibinfo{person}{Beidi Chen},
  \bibinfo{person}{Lili Yu}, \bibinfo{person}{Hao Zhang},
  \bibinfo{person}{Jonathan May}, \bibinfo{person}{Luke Zettlemoyer},
  \bibinfo{person}{Omer Levy}, {and} \bibinfo{person}{Chunting Zhou}.}
  \bibinfo{year}{2024}\natexlab{a}.
\newblock \bibinfo{title}{Megalodon: Efficient llm pretraining and inference
  with unlimited context length}.
\newblock \bibinfo{howpublished}{arXiv preprint arXiv:2404.08801}.
\newblock


\bibitem[\protect\citeauthoryear{Ma, Zhang, Zhang, Yu, Zhang, Zhang, Luo, Wang,
  and Tang}{Ma et~al\mbox{.}}{2024b}]%
        {ma2024spreadsheetbench}
\bibfield{author}{\bibinfo{person}{Zeyao Ma}, \bibinfo{person}{Bohan Zhang},
  \bibinfo{person}{Jing Zhang}, \bibinfo{person}{Jifan Yu},
  \bibinfo{person}{Xiaokang Zhang}, \bibinfo{person}{Xiaohan Zhang},
  \bibinfo{person}{Sijia Luo}, \bibinfo{person}{Xi Wang}, {and}
  \bibinfo{person}{Jie Tang}.} \bibinfo{year}{2024}\natexlab{b}.
\newblock \showarticletitle{SpreadsheetBench: Towards Challenging Real World
  Spreadsheet Manipulation}. In \bibinfo{booktitle}{\emph{The Thirty-eight
  Conference on Neural Information Processing Systems Datasets and Benchmarks
  Track}}.
\newblock
\urldef\tempurl%
\url{https://openreview.net/forum?id=KYxzmRLF6i}
\showURL{%
\tempurl}


\bibitem[\protect\citeauthoryear{McDermott, Arshad, Petyuk, Fu, Gritsenko,
  Clauss, Moore, Schepmoes, Zhao, Monroe, et~al\mbox{.}}{McDermott
  et~al\mbox{.}}{2020}]%
        {mcdermott2020proteogenomic}
\bibfield{author}{\bibinfo{person}{Jason~E McDermott}, \bibinfo{person}{Osama~A
  Arshad}, \bibinfo{person}{Vladislav~A Petyuk}, \bibinfo{person}{Yi Fu},
  \bibinfo{person}{Marina~A Gritsenko}, \bibinfo{person}{Therese~R Clauss},
  \bibinfo{person}{Ronald~J Moore}, \bibinfo{person}{Athena~A Schepmoes},
  \bibinfo{person}{Rui Zhao}, \bibinfo{person}{Matthew~E Monroe},
  {et~al\mbox{.}}} \bibinfo{year}{2020}\natexlab{}.
\newblock \showarticletitle{Proteogenomic characterization of ovarian HGSC
  implicates mitotic kinases, replication stress in observed chromosomal
  instability}.
\newblock \bibinfo{journal}{\emph{Cell reports medicine}} \bibinfo{volume}{1},
  \bibinfo{number}{1} (\bibinfo{year}{2020}).
\newblock


\bibitem[\protect\citeauthoryear{{Muñoz Monjas}, {Rubio Ruiz}, {Pérez del
  Rey}, and Palchuk}{{Muñoz Monjas} et~al\mbox{.}}{2025}]%
        {healthcareInteroperability2025}
\bibfield{author}{\bibinfo{person}{Aída {Muñoz Monjas}},
  \bibinfo{person}{David {Rubio Ruiz}}, \bibinfo{person}{David {Pérez del
  Rey}}, {and} \bibinfo{person}{Matvey~B. Palchuk}.}
  \bibinfo{year}{2025}\natexlab{}.
\newblock \showarticletitle{Enhancing real world data interoperability in
  healthcare: A methodological approach to laboratory unit harmonization}.
\newblock \bibinfo{journal}{\emph{International Journal of Medical
  Informatics}}  \bibinfo{volume}{193} (\bibinfo{year}{2025}),
  \bibinfo{pages}{105665}.
\newblock
\showISSN{1386-5056}
\urldef\tempurl%
\url{https://doi.org/10.1016/j.ijmedinf.2024.105665}
\showDOI{\tempurl}


\bibitem[\protect\citeauthoryear{Narayan, Chami, Orr, and R{\'{e}}}{Narayan
  et~al\mbox{.}}{2022}]%
        {narayan-vldb2022}
\bibfield{author}{\bibinfo{person}{Avanika Narayan}, \bibinfo{person}{Ines
  Chami}, \bibinfo{person}{Laurel~J. Orr}, {and} \bibinfo{person}{Christopher
  R{\'{e}}}.} \bibinfo{year}{2022}\natexlab{}.
\newblock \showarticletitle{Can Foundation Models Wrangle Your Data?}
\newblock \bibinfo{journal}{\emph{Proc. {VLDB} Endow.}} \bibinfo{volume}{16},
  \bibinfo{number}{4} (\bibinfo{year}{2022}), \bibinfo{pages}{738--746}.
\newblock


\bibitem[\protect\citeauthoryear{of~Automotive Engineers~(SAE)}{of~Automotive
  Engineers~(SAE)}{2018}]%
        {society2018taxonomy}
\bibfield{author}{\bibinfo{person}{Society of Automotive Engineers~(SAE)}.}
  \bibinfo{year}{2018}\natexlab{}.
\newblock \bibinfo{title}{Taxonomy and Definitions for Terms Related to Driving
  Automation Systems for On-Road Motor Vehicles (J3016\_201806)}.
\newblock
\newblock


\bibitem[\protect\citeauthoryear{Ray}{Ray}{[n.d.]}]%
        {zdnet-github-copilot}
\bibfield{author}{\bibinfo{person}{Tiernan Ray}.}
  \bibinfo{year}{[n.d.]}\natexlab{}.
\newblock \bibinfo{title}{Microsoft has over a million paying Github Copilot
  users: CEO Nadella}.
\newblock
  \bibinfo{howpublished}{\url{https://www.zdnet.com/article/microsoft-has-over-a-million-paying-github-copilot-users-ceo-nadella/}}.
\newblock


\bibitem[\protect\citeauthoryear{Rupprecht, Davis, Arnold, Gur, and
  Bhagwat}{Rupprecht et~al\mbox{.}}{2020}]%
        {rupprechtVLDB2020}
\bibfield{author}{\bibinfo{person}{Lukas Rupprecht}, \bibinfo{person}{James~C.
  Davis}, \bibinfo{person}{Constantine Arnold}, \bibinfo{person}{Yaniv Gur},
  {and} \bibinfo{person}{Deepavali Bhagwat}.} \bibinfo{year}{2020}\natexlab{}.
\newblock \showarticletitle{Improving reproducibility of data science pipelines
  through transparent provenance capture}.
\newblock \bibinfo{journal}{\emph{Proc. VLDB Endow.}} \bibinfo{volume}{13},
  \bibinfo{number}{12} (\bibinfo{date}{Aug.} \bibinfo{year}{2020}),
  \bibinfo{pages}{3354–3368}.
\newblock
\showISSN{2150-8097}
\urldef\tempurl%
\url{https://doi.org/10.14778/3415478.3415556}
\showDOI{\tempurl}


\bibitem[\protect\citeauthoryear{Russell and Norvig}{Russell and
  Norvig}{2016}]%
        {russell2016artificial}
\bibfield{author}{\bibinfo{person}{Stuart~J Russell} {and}
  \bibinfo{person}{Peter Norvig}.} \bibinfo{year}{2016}\natexlab{}.
\newblock \bibinfo{booktitle}{\emph{Artificial intelligence: a modern
  approach}}.
\newblock \bibinfo{publisher}{Pearson}.
\newblock


\bibitem[\protect\citeauthoryear{Santos, Castelo, Felix, Ono, Yu, Hong, Silva,
  Bertini, and Freire}{Santos et~al\mbox{.}}{2019}]%
        {santos2019visus}
\bibfield{author}{\bibinfo{person}{A{\'e}cio Santos}, \bibinfo{person}{Sonia
  Castelo}, \bibinfo{person}{Cristian Felix}, \bibinfo{person}{Jorge~Piazentin
  Ono}, \bibinfo{person}{Bowen Yu}, \bibinfo{person}{Sungsoo~Ray Hong},
  \bibinfo{person}{Cl{\'a}udio~T Silva}, \bibinfo{person}{Enrico Bertini},
  {and} \bibinfo{person}{Juliana Freire}.} \bibinfo{year}{2019}\natexlab{}.
\newblock \showarticletitle{Visus: An interactive system for automatic machine
  learning model building and curation}. In
  \bibinfo{booktitle}{\emph{Proceedings of the Workshop on Human-In-the-Loop
  Data Analytics}}. \bibinfo{pages}{1--7}.
\newblock


\bibitem[\protect\citeauthoryear{Satpathy, Krug, Beltran, Savage, Petralia,
  Kumar-Sinha, Dou, Reva, Kane, Avanessian, et~al\mbox{.}}{Satpathy
  et~al\mbox{.}}{2021}]%
        {satpathy2021proteogenomic}
\bibfield{author}{\bibinfo{person}{Shankha Satpathy}, \bibinfo{person}{Karsten
  Krug}, \bibinfo{person}{Pierre M~Jean Beltran}, \bibinfo{person}{Sara~R
  Savage}, \bibinfo{person}{Francesca Petralia}, \bibinfo{person}{Chandan
  Kumar-Sinha}, \bibinfo{person}{Yongchao Dou}, \bibinfo{person}{Boris Reva},
  \bibinfo{person}{M~Harry Kane}, \bibinfo{person}{Shayan~C Avanessian},
  {et~al\mbox{.}}} \bibinfo{year}{2021}\natexlab{}.
\newblock \showarticletitle{A proteogenomic portrait of lung squamous cell
  carcinoma}.
\newblock \bibinfo{journal}{\emph{Cell}} \bibinfo{volume}{184},
  \bibinfo{number}{16} (\bibinfo{year}{2021}), \bibinfo{pages}{4348--4371}.
\newblock


\bibitem[\protect\citeauthoryear{Shang, Zgraggen, Buratti, Kossmann, Eichmann,
  Chung, Binnig, Upfal, and Kraska}{Shang et~al\mbox{.}}{2019}]%
        {shang2019democratizing}
\bibfield{author}{\bibinfo{person}{Zeyuan Shang}, \bibinfo{person}{Emanuel
  Zgraggen}, \bibinfo{person}{Benedetto Buratti}, \bibinfo{person}{Ferdinand
  Kossmann}, \bibinfo{person}{Philipp Eichmann}, \bibinfo{person}{Yeounoh
  Chung}, \bibinfo{person}{Carsten Binnig}, \bibinfo{person}{Eli Upfal}, {and}
  \bibinfo{person}{Tim Kraska}.} \bibinfo{year}{2019}\natexlab{}.
\newblock \showarticletitle{Democratizing data science through interactive
  curation of ml pipelines}. In \bibinfo{booktitle}{\emph{Proceedings of the
  2019 international conference on management of data}}.
  \bibinfo{pages}{1171--1188}.
\newblock


\bibitem[\protect\citeauthoryear{sklearn-pipelines}{sklearn-pipelines}{[n.d.]}]%
        {sklearn-pipelines}
sklearn-pipelines \bibinfo{year}{[n.d.]}\natexlab{}.
\newblock \bibinfo{title}{scikit-learn: Pipelines and composite estimators}.
\newblock
  \bibinfo{howpublished}{\url{https://scikit-learn.org/1.6/modules/compose.html}}.
\newblock


\bibitem[\protect\citeauthoryear{Stureborg, Alikaniotis, and Suhara}{Stureborg
  et~al\mbox{.}}{2024}]%
        {stureborg2024large}
\bibfield{author}{\bibinfo{person}{Rickard Stureborg},
  \bibinfo{person}{Dimitris Alikaniotis}, {and} \bibinfo{person}{Yoshi
  Suhara}.} \bibinfo{year}{2024}\natexlab{}.
\newblock \showarticletitle{Large language models are inconsistent and biased
  evaluators}.
\newblock \bibinfo{journal}{\emph{arXiv preprint arXiv:2405.01724}}
  (\bibinfo{year}{2024}).
\newblock


\bibitem[\protect\citeauthoryear{Sui, Zhou, Zhou, Han, and Zhang}{Sui
  et~al\mbox{.}}{2024}]%
        {TableMeetsLLM2024}
\bibfield{author}{\bibinfo{person}{Yuan Sui}, \bibinfo{person}{Mengyu Zhou},
  \bibinfo{person}{Mingjie Zhou}, \bibinfo{person}{Shi Han}, {and}
  \bibinfo{person}{Dongmei Zhang}.} \bibinfo{year}{2024}\natexlab{}.
\newblock \showarticletitle{Table Meets LLM: Can Large Language Models
  Understand Structured Table Data? A Benchmark and Empirical Study}
  \emph{(\bibinfo{series}{WSDM '24})}. \bibinfo{publisher}{Association for
  Computing Machinery}, \bibinfo{address}{New York, NY, USA},
  \bibinfo{pages}{645–654}.
\newblock
\showISBNx{9798400703713}
\urldef\tempurl%
\url{https://doi.org/10.1145/3616855.3635752}
\showDOI{\tempurl}


\bibitem[\protect\citeauthoryear{Trummer}{Trummer}{2022}]%
        {trummer2022codexdb}
\bibfield{author}{\bibinfo{person}{Immanuel Trummer}.}
  \bibinfo{year}{2022}\natexlab{}.
\newblock \showarticletitle{CodexDB: Synthesizing code for query processing
  from natural language instructions using GPT-3 Codex}.
\newblock \bibinfo{journal}{\emph{Proceedings of the VLDB Endowment}}
  \bibinfo{volume}{15}, \bibinfo{number}{11} (\bibinfo{year}{2022}),
  \bibinfo{pages}{2921--2928}.
\newblock


\bibitem[\protect\citeauthoryear{Tu, Fan, Tang, Wang, Li, Du, Jia, and Gao}{Tu
  et~al\mbox{.}}{2023}]%
        {tu2023unicorn}
\bibfield{author}{\bibinfo{person}{Jianhong Tu}, \bibinfo{person}{Ju Fan},
  \bibinfo{person}{Nan Tang}, \bibinfo{person}{Peng Wang},
  \bibinfo{person}{Guoliang Li}, \bibinfo{person}{Xiaoyong Du},
  \bibinfo{person}{Xiaofeng Jia}, {and} \bibinfo{person}{Song Gao}.}
  \bibinfo{year}{2023}\natexlab{}.
\newblock \showarticletitle{Unicorn: A unified multi-tasking model for
  supporting matching tasks in data integration}.
\newblock \bibinfo{journal}{\emph{Proceedings of the ACM on Management of
  Data}} \bibinfo{volume}{1}, \bibinfo{number}{1} (\bibinfo{year}{2023}),
  \bibinfo{pages}{1--26}.
\newblock


\bibitem[\protect\citeauthoryear{Vasaikar, Huang, Wang, Petyuk, Savage, Wen,
  Dou, Zhang, Shi, Arshad, et~al\mbox{.}}{Vasaikar et~al\mbox{.}}{2019}]%
        {vasaikar2019proteogenomic}
\bibfield{author}{\bibinfo{person}{Suhas Vasaikar}, \bibinfo{person}{Chen
  Huang}, \bibinfo{person}{Xiaojing Wang}, \bibinfo{person}{Vladislav~A
  Petyuk}, \bibinfo{person}{Sara~R Savage}, \bibinfo{person}{Bo Wen},
  \bibinfo{person}{Yongchao Dou}, \bibinfo{person}{Yun Zhang},
  \bibinfo{person}{Zhiao Shi}, \bibinfo{person}{Osama~A Arshad},
  {et~al\mbox{.}}} \bibinfo{year}{2019}\natexlab{}.
\newblock \showarticletitle{Proteogenomic analysis of human colon cancer
  reveals new therapeutic opportunities}.
\newblock \bibinfo{journal}{\emph{Cell}} \bibinfo{volume}{177},
  \bibinfo{number}{4} (\bibinfo{year}{2019}), \bibinfo{pages}{1035--1049}.
\newblock


\bibitem[\protect\citeauthoryear{Wang, Li, and Hirota}{Wang
  et~al\mbox{.}}{2021b}]%
        {wang2021machamp}
\bibfield{author}{\bibinfo{person}{Jin Wang}, \bibinfo{person}{Yuliang Li},
  {and} \bibinfo{person}{Wataru Hirota}.} \bibinfo{year}{2021}\natexlab{b}.
\newblock \showarticletitle{Machamp: A generalized entity matching benchmark}.
  In \bibinfo{booktitle}{\emph{Proceedings of the 30th ACM International
  Conference on Information \& Knowledge Management}}.
  \bibinfo{pages}{4633--4642}.
\newblock


\bibitem[\protect\citeauthoryear{Wang, Ma, Feng, Zhang, Yang, Zhang, Chen,
  Tang, Chen, Lin, et~al\mbox{.}}{Wang et~al\mbox{.}}{2024a}]%
        {wang2024-agents-survey}
\bibfield{author}{\bibinfo{person}{Lei Wang}, \bibinfo{person}{Chen Ma},
  \bibinfo{person}{Xueyang Feng}, \bibinfo{person}{Zeyu Zhang},
  \bibinfo{person}{Hao Yang}, \bibinfo{person}{Jingsen Zhang},
  \bibinfo{person}{Zhiyuan Chen}, \bibinfo{person}{Jiakai Tang},
  \bibinfo{person}{Xu Chen}, \bibinfo{person}{Yankai Lin}, {et~al\mbox{.}}}
  \bibinfo{year}{2024}\natexlab{a}.
\newblock \showarticletitle{A survey on large language model based autonomous
  agents}.
\newblock \bibinfo{journal}{\emph{Frontiers of Computer Science}}
  \bibinfo{volume}{18}, \bibinfo{number}{6} (\bibinfo{year}{2024}),
  \bibinfo{pages}{186345}.
\newblock


\bibitem[\protect\citeauthoryear{Wang, Karpova, Gritsenko, Kyle, Cao, Li,
  Rykunov, Colaprico, Rothstein, Hong, et~al\mbox{.}}{Wang
  et~al\mbox{.}}{2021a}]%
        {wang2021proteogenomic}
\bibfield{author}{\bibinfo{person}{Liang-Bo Wang}, \bibinfo{person}{Alla
  Karpova}, \bibinfo{person}{Marina~A Gritsenko}, \bibinfo{person}{Jennifer~E
  Kyle}, \bibinfo{person}{Song Cao}, \bibinfo{person}{Yize Li},
  \bibinfo{person}{Dmitry Rykunov}, \bibinfo{person}{Antonio Colaprico},
  \bibinfo{person}{Joseph~H Rothstein}, \bibinfo{person}{Runyu Hong},
  {et~al\mbox{.}}} \bibinfo{year}{2021}\natexlab{a}.
\newblock \showarticletitle{Proteogenomic and metabolomic characterization of
  human glioblastoma}.
\newblock \bibinfo{journal}{\emph{Cancer cell}} \bibinfo{volume}{39},
  \bibinfo{number}{4} (\bibinfo{year}{2021}), \bibinfo{pages}{509--528}.
\newblock


\bibitem[\protect\citeauthoryear{Wang, Haas, and Meliou}{Wang
  et~al\mbox{.}}{2018}]%
        {WangHM2018}
\bibfield{author}{\bibinfo{person}{Xiaolan Wang}, \bibinfo{person}{Laura Haas},
  {and} \bibinfo{person}{Alexandra Meliou}.} \bibinfo{year}{2018}\natexlab{}.
\newblock
  \showarticletitle{\href{http://sites.computer.org/debull/A18june/p47.pdf}{Explaining
  Data Integration}}.
\newblock \bibinfo{journal}{\emph{IEEE Data Engineering Bulletin}}
  \bibinfo{volume}{41}, \bibinfo{number}{2} (\bibinfo{date}{June}
  \bibinfo{year}{2018}), \bibinfo{pages}{47--58}.
\newblock


\bibitem[\protect\citeauthoryear{Wang, Zhang, Li, Eisenschlos, Perot, Wang,
  Miculicich, Fujii, Shang, Lee, and Pfister}{Wang et~al\mbox{.}}{2024b}]%
        {wang2024chainoftable}
\bibfield{author}{\bibinfo{person}{Zilong Wang}, \bibinfo{person}{Hao Zhang},
  \bibinfo{person}{Chun-Liang Li}, \bibinfo{person}{Julian~Martin Eisenschlos},
  \bibinfo{person}{Vincent Perot}, \bibinfo{person}{Zifeng Wang},
  \bibinfo{person}{Lesly Miculicich}, \bibinfo{person}{Yasuhisa Fujii},
  \bibinfo{person}{Jingbo Shang}, \bibinfo{person}{Chen-Yu Lee}, {and}
  \bibinfo{person}{Tomas Pfister}.} \bibinfo{year}{2024}\natexlab{b}.
\newblock \showarticletitle{Chain-of-Table: Evolving Tables in the Reasoning
  Chain for Table Understanding}. In \bibinfo{booktitle}{\emph{The Twelfth
  International Conference on Learning Representations}}.
\newblock
\urldef\tempurl%
\url{https://openreview.net/forum?id=4L0xnS4GQM}
\showURL{%
\tempurl}


\bibitem[\protect\citeauthoryear{Wei, Wang, Schuurmans, Bosma, Xia, Chi, Le,
  Zhou, et~al\mbox{.}}{Wei et~al\mbox{.}}{2022}]%
        {wei2022chain}
\bibfield{author}{\bibinfo{person}{Jason Wei}, \bibinfo{person}{Xuezhi Wang},
  \bibinfo{person}{Dale Schuurmans}, \bibinfo{person}{Maarten Bosma},
  \bibinfo{person}{Fei Xia}, \bibinfo{person}{Ed Chi}, \bibinfo{person}{Quoc~V
  Le}, \bibinfo{person}{Denny Zhou}, {et~al\mbox{.}}}
  \bibinfo{year}{2022}\natexlab{}.
\newblock \showarticletitle{Chain-of-thought prompting elicits reasoning in
  large language models}.
\newblock \bibinfo{journal}{\emph{Advances in neural information processing
  systems}}  \bibinfo{volume}{35} (\bibinfo{year}{2022}),
  \bibinfo{pages}{24824--24837}.
\newblock


\bibitem[\protect\citeauthoryear{Wu, Bansal, Zhang, Wu, Zhang, Zhu, Li, Jiang,
  Zhang, and Wang}{Wu et~al\mbox{.}}{2023}]%
        {wu2023autogen}
\bibfield{author}{\bibinfo{person}{Qingyun Wu}, \bibinfo{person}{Gagan Bansal},
  \bibinfo{person}{Jieyu Zhang}, \bibinfo{person}{Yiran Wu},
  \bibinfo{person}{Shaokun Zhang}, \bibinfo{person}{Erkang Zhu},
  \bibinfo{person}{Beibin Li}, \bibinfo{person}{Li Jiang},
  \bibinfo{person}{Xiaoyun Zhang}, {and} \bibinfo{person}{Chi Wang}.}
  \bibinfo{year}{2023}\natexlab{}.
\newblock \showarticletitle{Autogen: Enabling next-gen llm applications via
  multi-agent conversation framework}.
\newblock \bibinfo{journal}{\emph{arXiv preprint arXiv:2308.08155}}
  (\bibinfo{year}{2023}).
\newblock


\bibitem[\protect\citeauthoryear{Xi, Chen, Guo, He, Ding, Hong, Zhang, Wang,
  Jin, Zhou, et~al\mbox{.}}{Xi et~al\mbox{.}}{2023}]%
        {xi2023-agents-survey}
\bibfield{author}{\bibinfo{person}{Zhiheng Xi}, \bibinfo{person}{Wenxiang
  Chen}, \bibinfo{person}{Xin Guo}, \bibinfo{person}{Wei He},
  \bibinfo{person}{Yiwen Ding}, \bibinfo{person}{Boyang Hong},
  \bibinfo{person}{Ming Zhang}, \bibinfo{person}{Junzhe Wang},
  \bibinfo{person}{Senjie Jin}, \bibinfo{person}{Enyu Zhou}, {et~al\mbox{.}}}
  \bibinfo{year}{2023}\natexlab{}.
\newblock \showarticletitle{The rise and potential of large language model
  based agents: A survey}.
\newblock \bibinfo{journal}{\emph{arXiv preprint arXiv:2309.07864}}
  (\bibinfo{year}{2023}).
\newblock


\bibitem[\protect\citeauthoryear{Yao, Zhao, Yu, Shafran, Narasimhan, and
  Cao}{Yao et~al\mbox{.}}{2022}]%
        {yao2022react}
\bibfield{author}{\bibinfo{person}{Shunyu Yao}, \bibinfo{person}{Jeffrey Zhao},
  \bibinfo{person}{Dian Yu}, \bibinfo{person}{Izhak Shafran},
  \bibinfo{person}{Karthik~R Narasimhan}, {and} \bibinfo{person}{Yuan Cao}.}
  \bibinfo{year}{2022}\natexlab{}.
\newblock \showarticletitle{ReAct: Synergizing Reasoning and Acting in Language
  Models}. In \bibinfo{booktitle}{\emph{NeurIPS 2022 Foundation Models for
  Decision Making Workshop}}.
\newblock
\urldef\tempurl%
\url{https://openreview.net/forum?id=tvI4u1ylcqs}
\showURL{%
\tempurl}


\bibitem[\protect\citeauthoryear{Yu and Silva}{Yu and Silva}{2019}]%
        {yu2019flowsense}
\bibfield{author}{\bibinfo{person}{Bowen Yu} {and}
  \bibinfo{person}{Cl{\'a}udio~T Silva}.} \bibinfo{year}{2019}\natexlab{}.
\newblock \showarticletitle{FlowSense: A natural language interface for visual
  data exploration within a dataflow system}.
\newblock \bibinfo{journal}{\emph{IEEE transactions on visualization and
  computer graphics}} \bibinfo{volume}{26}, \bibinfo{number}{1}
  (\bibinfo{year}{2019}), \bibinfo{pages}{1--11}.
\newblock


\bibitem[\protect\citeauthoryear{Zhang and Choi}{Zhang and Choi}{2023}]%
        {zhang2023clarify}
\bibfield{author}{\bibinfo{person}{Michael~JQ Zhang} {and}
  \bibinfo{person}{Eunsol Choi}.} \bibinfo{year}{2023}\natexlab{}.
\newblock \showarticletitle{Clarify when necessary: Resolving ambiguity through
  interaction with lms}.
\newblock \bibinfo{journal}{\emph{arXiv preprint arXiv:2311.09469}}
  (\bibinfo{year}{2023}).
\newblock


\bibitem[\protect\citeauthoryear{Zhang, Jiang, Han, Chen, Yang, and Ren}{Zhang
  et~al\mbox{.}}{2024}]%
        {zhang2024benchmarking}
\bibfield{author}{\bibinfo{person}{Yuge Zhang}, \bibinfo{person}{Qiyang Jiang},
  \bibinfo{person}{Xingyu Han}, \bibinfo{person}{Nan Chen},
  \bibinfo{person}{Yuqing Yang}, {and} \bibinfo{person}{Kan Ren}.}
  \bibinfo{year}{2024}\natexlab{}.
\newblock \showarticletitle{Benchmarking Data Science Agents}.
\newblock \bibinfo{journal}{\emph{arXiv preprint arXiv:2402.17168}}
  (\bibinfo{year}{2024}).
\newblock


\bibitem[\protect\citeauthoryear{Zhao, Chen, Yang, Liu, Deng, Cai, Wang, Yin,
  and Du}{Zhao et~al\mbox{.}}{2024}]%
        {explainability2024}
\bibfield{author}{\bibinfo{person}{Haiyan Zhao}, \bibinfo{person}{Hanjie Chen},
  \bibinfo{person}{Fan Yang}, \bibinfo{person}{Ninghao Liu},
  \bibinfo{person}{Huiqi Deng}, \bibinfo{person}{Hengyi Cai},
  \bibinfo{person}{Shuaiqiang Wang}, \bibinfo{person}{Dawei Yin}, {and}
  \bibinfo{person}{Mengnan Du}.} \bibinfo{year}{2024}\natexlab{}.
\newblock \showarticletitle{Explainability for Large Language Models: A
  Survey}.
\newblock \bibinfo{journal}{\emph{ACM Trans. Intell. Syst. Technol.}}
  \bibinfo{volume}{15}, \bibinfo{number}{2}, Article \bibinfo{articleno}{20}
  (\bibinfo{date}{Feb.} \bibinfo{year}{2024}), \bibinfo{numpages}{38}~pages.
\newblock
\showISSN{2157-6904}
\urldef\tempurl%
\url{https://doi.org/10.1145/3639372}
\showDOI{\tempurl}


\end{thebibliography}
% \balance

\end{document}